\documentclass{article}
\usepackage{amssymb}

\usepackage[final]{corl_2025} 

\usepackage{subfig}
\usepackage{wrapfig}
\usepackage{times}
\usepackage{epsfig}
\usepackage{graphicx}
\usepackage{amsmath}
\usepackage{csquotes}
\usepackage{amssymb}
\usepackage{algorithm}
\usepackage{pifont}
\usepackage{algorithmic}
\usepackage{mathrsfs}
\usepackage{booktabs, multirow} 

\usepackage{caption}
\definecolor{darkred}{RGB}{139, 0, 0}    
\definecolor{darkgreen}{RGB}{0, 100, 0}  

\newcommand{\se}{\operatorname{SE}}
\newcommand{\so}{\operatorname{SO}}

\newcommand{\ray}{\operatorname{ray}}

\newcommand{\R}{\mathbb{R}}

\DeclareRobustCommand\sbseries{\fontseries{sb}\selectfont}
\DeclareTextFontCommand{\textsb}{\sbseries}

\title{Motion-Blender Gaussian Splatting for \\ Dynamic Scene Reconstruction}
\author{Xinyu Zhang 
\qquad 
Haonan Chang
\qquad  Yuhan Liu
\qquad  Abdeslam Boularias \\
\{xz653, hc856, yl1834, ab1544\}@rutgers.edu \\
Rutgers University
}

\begin{document}
\maketitle



\begin{abstract} Gaussian splatting has emerged as a powerful tool for high-fidelity reconstruction of dynamic scenes.  However, existing methods primarily rely on implicit motion representations, such as encoding motions into neural networks or per-Gaussian parameters, which makes it difficult to further manipulate the reconstructed motions.
This lack of explicit controllability limits existing methods to replaying recorded motions only, which hinders a wider application in robotics.
To address this, we propose Motion Blender Gaussian Splatting (MBGS), a novel framework that uses motion graphs as an explicit and sparse motion representation. The motion of a graph's links is propagated to individual Gaussians via dual quaternion skinning, with learnable weight painting functions that determine the influence of each link. The motion graphs and 3D Gaussians are jointly optimized from input videos via differentiable rendering.
Experiments show that MBGS achieves state-of-the-art performance on the highly challenging {\it iPhone} dataset while being competitive on {\it HyperNeRF}. 
We demonstrate the application potential of our method in animating novel object poses, synthesizing real robot demonstrations, and predicting robot actions through visual planning.
The source code and the models are included in the supplementary material.
Video demonstrations can be found at
\href{https://mlzxy.github.io/motion-blender-gs/}{\texttt{mlzxy.github.io/motion-blender-gs}}. 


\end{abstract}


\section{Introduction}

Reconstructing and modeling dynamic 3D scenes is a fundamental challenge in robot vision.
Recent work on 3D Gaussian splatting has made significant progress in capturing dynamic scenes, enabling efficient and high-fidelity reconstruction~\cite{wang2024shape, kerbl20233d, wu20244d, li2024spacetime, yang2023real, xie2024physgaussian, Huang_2024_CVPR}.
Existing Gaussian splatting methods utilize 3D Gaussians to represent geometry and appearance, paired with motion modules that determine the movements of the Gaussians. For instance, 4D-Gaussians~\cite{wu20244d} and Deformable-GS~\cite{Yang2023Deformable3G} encode motion implicitly into neural networks. Shape-of-Motion~\cite{wang2024shape} and STG~\cite{li2024spacetime} use shallower models that require dense per-Gaussian motion parameters. 
While these approaches achieve high-fidelity reconstruction, the reconstructed scenes cannot be easily altered or manipulated in simulation due to their implicit motion representation, which limits their use for robot manipulation planning.
This lack of direct and explicit controllability restricts existing methods to simply replaying recorded motions.

Therefore, a key question to answer is: can we develop an explicit and sparse motion representation without compromising the ability to reconstruct complex dynamic scenes? To answer this question, we revisit in this work some classical animation techniques~\cite{james2005skinning}. Explicit motion representations, such as deformation graphs~\cite{sumner2007embedded}, harmonic coordinates~\cite{joshi2007harmonic}, and cage~\cite{nieto2012cage},  have been developed to animate objects with diverse motions. However, these classical hand-crafted methods focus on applying manually designed motions to mesh surfaces instead of differentiable models such as 3D Gaussians~\cite{blender}. Further, these methods are not able to reconstruct motion or geometry from images.

\begin{figure*}
\vspace{-0.5em}
\centering
\includegraphics[width=\linewidth]{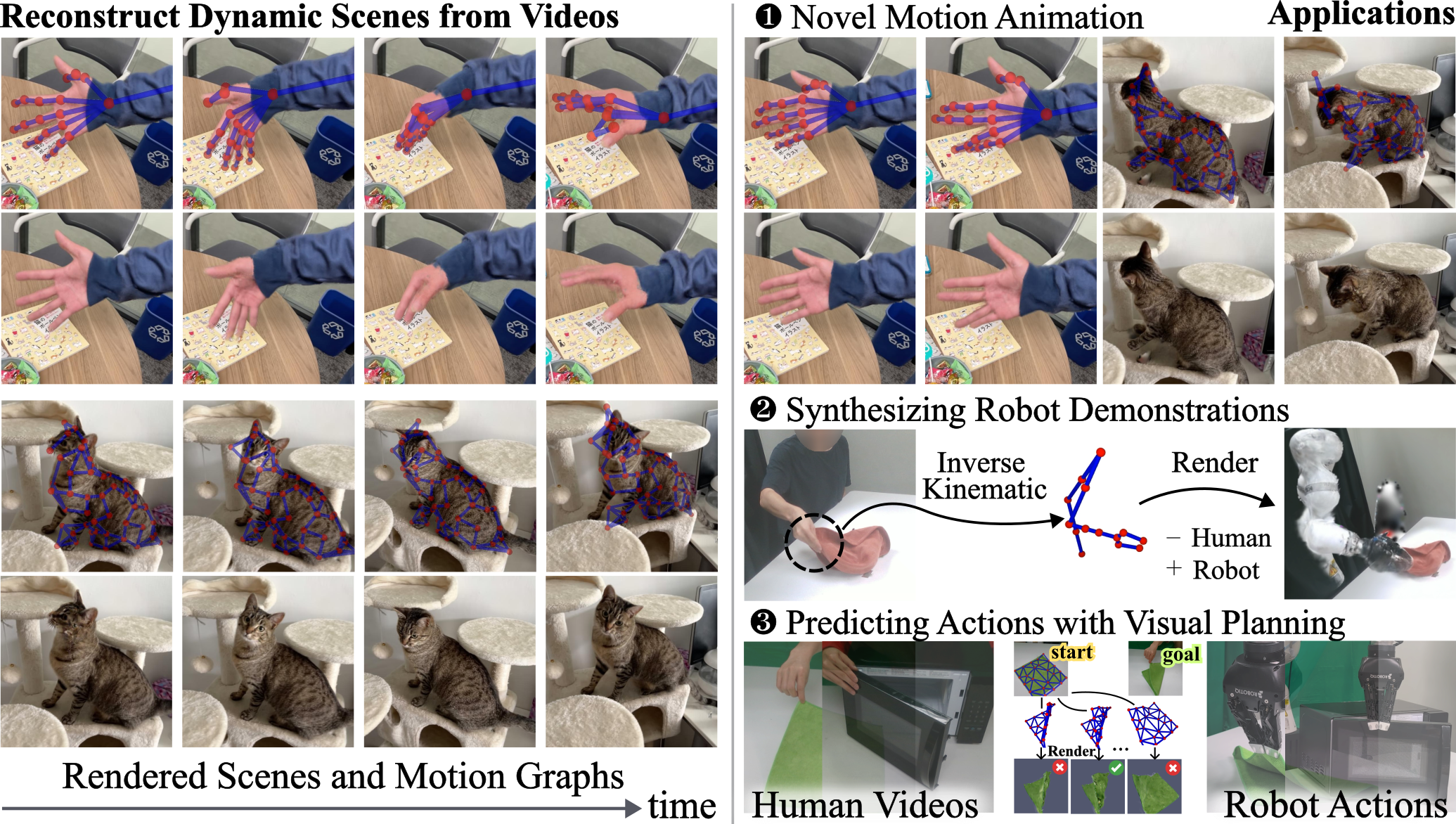}
\label{fig:teaser}
\caption{\textbf{Capabilities of Our Framework.} Our method reconstructs and renders dynamic scenes into 3D Gaussians and motion graphs from input videos. The learned motion graphs for a hand and cat are shown with their corresponding rendered scenes (left). Our approach enables three key applications (right): \ding{202} Novel pose animation through motion graph editing, \ding{203} Robot demonstration synthesis by using robot kinematic chains as motion graphs, and \ding{204} Predicting robot actions by simulating graph movements to minimize the difference between rendered and goal images. } 


\vspace{-0.5em}
\end{figure*}

Inspired by classical mesh animation techniques, we propose using kinematic trees and deformable graphs as motion representations for Gaussian splatting-based reconstruction. Kinematic trees are well-suited for capturing the motions of articulated objects, such as human bodies or robots, while deformable graphs, free from topological constraints, are ideal for modeling non-rigid object deformations. We collectively refer to both kinematic trees and deformable graphs as {\it motion graphs}.
Motions of graph links are propagated to individual Gaussians through dual quaternion skinning~\cite{kavan2007skinning}. The influence of each graph link on a Gaussian is predicted by a learnable weight painting function. 
The graphs are initialized from point clouds and 2D keypoints at a canonical frame. 
Both the Gaussians and motion graphs are optimized end-to-end jointly from videos via differentiable rendering. 
We term our approach Motion-Blender Gaussian Splatting (MBGS).

Our main contributions can be summarized as follows.
(1) We introduce a new dynamic Gaussian splatting framework based on explicit and sparse motion graphs, which allows for straightforward robot manipulation planning in  reconstructed scenes. Gaussian motions are predicted by blending the  link motions. (2) We propose two types of parametric motion graphs---kinematic trees and deformable graphs---and a learnable weight painting function based on Gaussian kernels, along with optimization details in Sec.~\ref{sec:optimization}. Our method learns both 3D Gaussians and motion graphs jointly. (3) Compared with state-of-the-art, our method outperforms Shape-of-Motion on the challenging iPhone dataset~\cite{gao2022monocular}, and achieves competitive performance with 4DGaussians on HyperNerf~\cite{park2021hypernerf}. Further, we demonstrate the applications of our approach on animating novel object motions, synthesizing real robot demonstrations, and predicting robot actions through visual planning. This can lead to significant efficiency improvement in gathering training datasets for robot learning~\cite{pmlr-v229-walke23a}.

\begin{figure*}[t]
    \vspace{-0.5em}
    \centering
    \includegraphics[width=\linewidth]{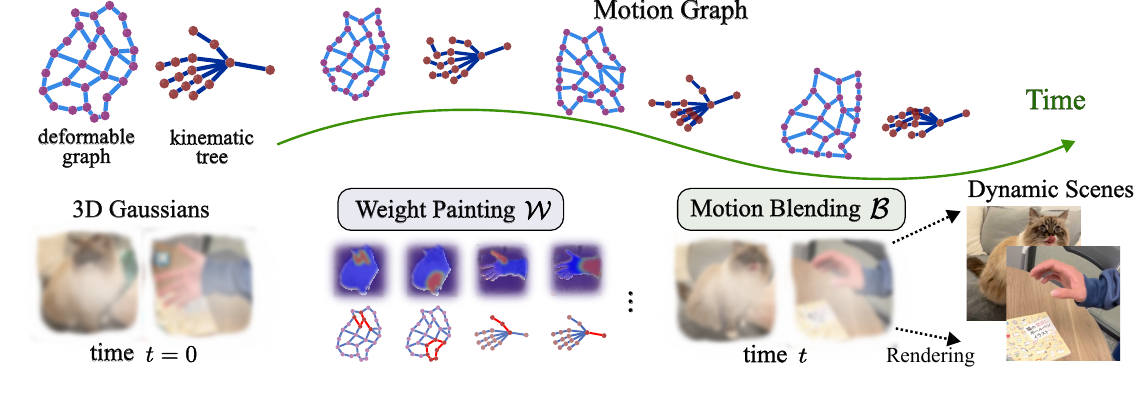}
    \caption{\textbf{Motion Blender Gaussian Splatting.} Our framework explicitly represents motion using sparse dynamic graphs. Static 3D Gaussians are associated with the graphs through learnable weight painting. Then, link-wise motions are propagated to the Gaussians through motion blending with dual quaternion skinning. We employ two motion graph types: kinematic trees, ideal for capturing articulated structures like human bodies, and deformable graphs, designed for modeling non-rigid deformations in soft objects. The parameters of the motion graph, weight painting functions, and 3D Gaussians are jointly optimized, end-to-end, via differentiable rendering.
    }
    \label{fig:framework}
    \vspace{-0.5em}
\end{figure*}

\section{Method}
\label{sec:m}
\vspace{-1em}


\vspace{0.5em}\noindent\textbf{Preliminaries.} Gaussian splatting represents a static 3D scene explicitly with a set  $\mathcal{G}$ of 3D Gaussians. Each 3D Gaussian $g \in \mathcal{G}$ is parameterized by its pose $\mathbf{p} \in \se(3)$, scale $\mathbf{s} \in \mathbb{R}^3$, opacity $o \in \mathbb{R}$ and color $\mathbf{c} \in \mathbb{R}^3$. 2D images can be rendered from $\mathcal{G}$ by blending the colors of overlapping Gaussians along the ray direction of each pixel. 
Dynamic Gaussian splatting extends each Gaussian by making it time-dependent, $g_t$, where $t\in [0,T-1]$ and $T$ represents the number of frames in the source video. The common practice is to make the pose time-dependent, $\mathbf{p}_t$, and keep the rest static. $\mathbf{p}_t$ is often obtained from a dynamic model $f$ as in Eq.~\ref{eq:pt1}, where $\theta$ denotes the model's parameters. 

\vspace{-1em}
\begin{equation}
\mathbf{p}_t = f(\mathbf{p}_0, t; \theta), \text{where}\,\, t > 0
\label{eq:pt1}
\end{equation}

$f$ is often implemented as a neural net, \emph{e.g.}, a deformation field network~\cite{wu20244d, Yang2023Deformable3G}, or as a shallow model such as motion coefficients or polynomials~\cite{lin2024gaussian, li2024spacetime, wang2024shape} that require per-Gaussian motion parameters.

\subsection{Motion Blender Gaussian Splatting}
\label{sec:m:framework}

Our framework is designed based on two key ideas that set it apart from existing works. \boxed{1} Represent motions with a sparse structure: Instead of associating dense motion parameters to each Gaussian (often numbering in the hundreds of thousands), we use a motion graph with far fewer parameters (often less than 100). The sparse link-wise motions are then propagated to each Gaussian through motion blending.
\boxed{2} Represent motions explicitly: Instead of encoding motions implicitly into neural networks, our motion graph is an explicit kinematic model. This allows a straightforward visualization and manipulation of motions in the 3D space, such as transferring motion patterns from one object to a new object, or creating novel animations through motion graph editing.


Formally, we define the motion graph $G_t = (\mathcal{J}_t, \mathcal{L})$ at time $t$ as a set of joints $\mathcal{J}_t$ with a set of edges $\mathcal{L}$, where $\max(|\mathcal{J}_t|, |\mathcal{L}|) \ll |\mathcal{G}|$. Each link $l \in \mathcal{L}$ is defined as $l = (s_l, e_l)$ and consists of the start and end joint indexes $s_l, e_l \in [1,|\mathcal{J}_t|]$. 
The positions of all the links at time $t$ are arranged as a matrix $X_{\mathcal{L},t} \in \mathbb{R}^{|\mathcal{L}| \times 6}$, where each row corresponds to the 3D coordinates of the start and end joints for a link in $\mathcal{L}$. Similarly, $\mathbf{x}_{t} \in \R^3$ denotes the position of a Gaussian with pose $\mathbf{p}_{t}$ at time $t$.
We assume both the edges $\mathcal{L}$ and the number of joints $|\mathcal{J}_t|$ stay static, and only joint positions undergo motions over time. 
Let $P_{\mathcal{L},t} \in \se(3)^{|\mathcal{L}|}$ be the poses of all the links at time $t$. Let $\mathbf{p}_0$ be the parameter that describes the initial pose of a Gaussian, then $\mathbf{p}_t$ is given by Eq.~\ref{eq:framework}.

\vspace{-0.25em}

\begin{equation}
\mathbf{p}_t = \mathcal{B}\big(\mathcal{R}\left(P_{\mathcal{L},0}, P_{\mathcal{L},t}\right), \mathcal{W}\left(\mathbf{x}_{0}, X_{\mathcal{L},0}\right)\big) \cdot \mathbf{p}_0
\label{eq:framework}    
\end{equation}

$\mathcal{W}: \R^3 \times \mathbb{R}^{|\mathcal{L}|\times6} \mapsto \Delta^{|\mathcal{L}|-1}$ denotes the \textit{weight painting function}, where $\Delta^{|\mathcal{L}|-1}$ represents the probability vector space defined as $\{ \mathbf{w} \in \mathbb{R}_+^{|\mathcal{L}|} \ | \ \sum_{i=1}^{|\mathcal{L}|} \mathbf{w}_i = 1\}$. This function $\mathcal{W}$ estimates the level of influence of each link $l\in \mathcal{L}$ on each Gaussian based on their relative positions at $t=0$.

$\mathcal{R}: \se(3)^{|\mathcal{L}|} \times \se(3)^{|\mathcal{L}|} \mapsto \se(3)^{|\mathcal{L}|}$ returns  relative rigid transforms between two sets of 3D poses. Specifically, $\mathcal{R}\left(P_{\mathcal{L},0}, P_{\mathcal{L},t}\right)$ indicates the SE(3) movement from time $0$ to time $t$ of each link in $\mathcal{L}$.

$\mathcal{B}: \se(3)^{|\mathcal{L}|} \times \Delta^{|\mathcal{L}|-1} \mapsto \se(3)$ represents the \textit{motion blending function}, which computes the per-Gaussian motion by propagating the link-wise movements $\mathcal{R}\left(P_{\mathcal{L},0}, P_{\mathcal{L},t}\right)$ to each Gaussian based on the weights assigned by $\mathcal{W}$ at $t=0$.

We implement $\mathcal{B}$ using dual quaternion skinning (DQS)~\cite{kavan2007skinning}. DQS represents transformations using dual quaternions and computes weighted averages in a way that guarantees the resulting transformation remains valid in SE(3) space. The link-wise movements computed by $\mathcal{R}$ are analytically determined, as the relative SE(3) transforms can be straightforwardly derived. Therefore, our motion blender framework hinges on two key design choices:  (1) Modeling graph motions by representing their temporal evolution through link poses $P_{\mathcal{L},t}$.  (2) Defining a weight painting function $\mathcal{W}$ that accurately captures the influence of each graph link on a Gaussian’s motion. We address the former by presenting two types of parametric motion graphs in Sec.~\ref{sec:m:motiongraph}. The latter, the design of $\mathcal{W}$, is detailed in Sec.~\ref{sec:m:weipaint}. The overall motion-blender Gaussian splatting framework is illustrated in Fig.~\ref{fig:framework}.

\subsection{Motion Graph Representation}
\label{sec:m:motiongraph}
\vspace{-0.5em}

\begin{figure}[!h]
    \centering
    \includegraphics[width=\linewidth]{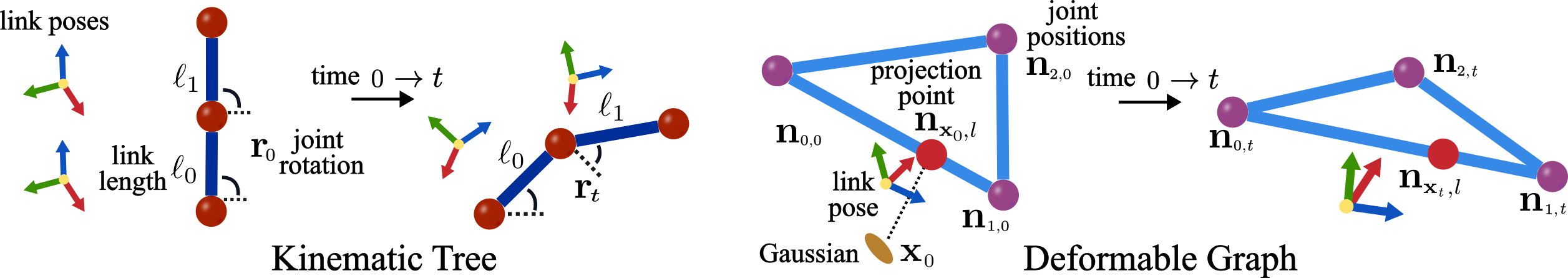}
    \caption{\textbf{Motion Graphs.} A kinematic tree (left) uses time-independent link lengths $\ell$ and dynamic joint rotations $\mathbf{r}_t \in \so(3)$. 
    Link poses (shown as colored coordinate axes) in world coordinates are computed through forward kinematics. A deformable graph (right) employs free-form topology parameterized by joint positions $\{\mathbf{n}_{i,t}\}$ and has non-rigid link deformations. Rigid per-link poses are obtained relative to each Gaussian position $\mathbf{x}_0$ and look-at transformation as in Eq.~\ref{eq:project-point} and Eq.~\ref{eq:lookat}.}
    \label{fig:motion-graph}
\end{figure}

To predict $P_{\mathcal{L},t}$, the link poses at time $t$, we use a parameterized function $\mathcal{P_L}(\theta, \phi) \in \se(3)^{|\mathcal{L}|}$.
Here, $\theta$ is the graph parameters that determine the link poses at a given instant, and $\phi$ is the time-independent graph parameters. By modeling $\theta$ as a time-varying sequence $(\theta_t)_{t=0}^{T-1}$, the link poses at time $t$ are defined as $P_{\mathcal{L},t} = \mathcal{P_L}(\theta_t, \phi)$, where $\theta_t$ describes the kinematic state of the graph at time $t$. Two types of motion graphs, kinematic trees and deformable graph, are shown in Fig.~\ref{fig:motion-graph}.

\noindent\textbf{Kinematic Tree.} A kinematic tree is a hierarchical, acyclic graph with a root joint. This representation is ideal for capturing the kinematic chains of articulated objects such as human bodies, robot arms, or other multi-joint systems. 
We parameterize the tree using $\theta = (\mathbf{r}, \mathbf{X}), \phi = (\ell_i)_{i=1}^L$, where $\ell_i \in \mathbb{R}^+$ denotes the length of the $i$-th link, $\mathbf{r} \in \so(3)^{|\mathcal{J}_t|-1}$ represents the joint rotations, 
and $\mathbf{X} \in \se(3)$ denotes the pose of the root joint. Then $\mathcal{P_L}(\theta_t, \phi)$ can be implemented via a forward kinematics algorithm~\cite{kucuk2006robot}, which propagates transformations from the root node through the tree, compounding rotations and translations across joints to derive every link pose. The forward kinematic procedure is differentiable, which enables the learning of $\theta$ and $\phi$ through back-propagation.

\begin{figure*}[t]
    \vspace{-1.5em}
    \centering
    \includegraphics[width=\linewidth]{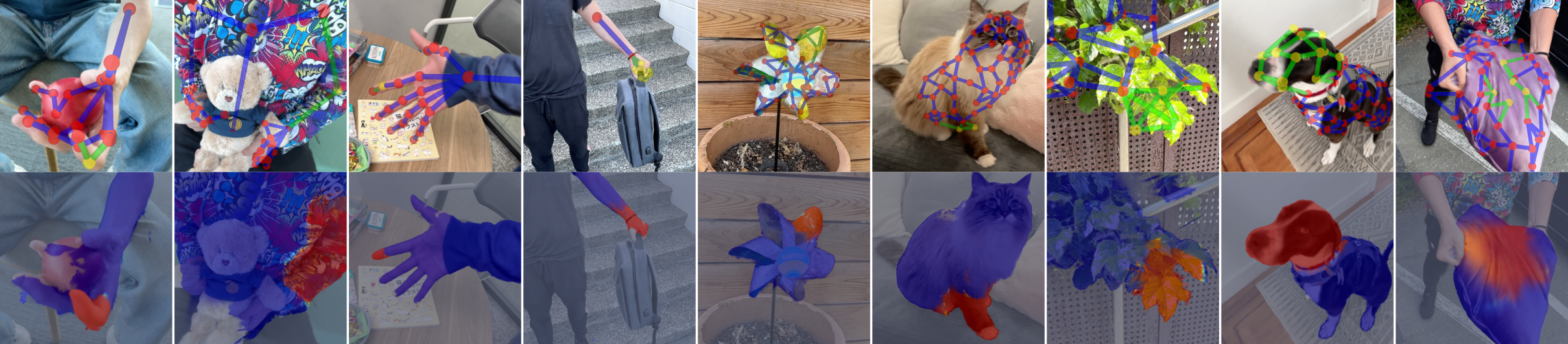}
    \caption{\textbf{Learned Motion Graphs and Weight Paintings.} The first row overlays the learned motion graphs on the images. The second row shows  painted weights (\textcolor{darkred}{red}) of graph links (\textcolor{darkgreen}{green}).}
    \label{fig:motion-graph-demo}
    \vspace{-1.5em}
\end{figure*}

\noindent\textbf{Deformable Graph.} Unlike the kinematic tree, a deformable graph imposes no topological constraints, allowing the joints to move freely in the 3D space. This representation is ideal for modeling non-rigid deformations and surface variations in soft objects.
We parameterize the graph as $\theta = (\mathbf{n}_j \in \mathbb{R}^3)_{j=1}^{|\mathcal{J}_t|}$, where $\mathbf{n}_j$ denotes the 3D position of the $j$-th joint.
However, the absence of topological constraints allows links between joints to stretch or compress, making it impossible to describe link poses using rigid transformations. 
To address this, we project each Gaussian $\mathbf{x}$ in the first frame on each link in the graph, and track the positions of the projected points in the remaining frames. Given each Gaussian $\mathbf{x}$, the matrix of the poses of all the links, denoted as $P_{\mathcal{L}}(\theta, \mathbf{x})$, is given by the 3D positions of the closest-point projections of $\mathbf{x}$ on the links in addition to the 3D directions of the links.
Specifically, for each link $l$ with start joint $\mathbf{n}_{s_l,0}$ and end joint $\mathbf{n}_{e_l,0}$ at $t=0$, we define $\mathbf{n}_{\mathbf{x}_{0},l} \in \mathbb{R}^3$ as the point on $l$ that has the minimal distance to $\mathbf{x}_{0}$. We refer $\mathbf{n}_{\mathbf{x}_{0},l}$ as the projection point at $t=0$.
Next, we define the projection point $\mathbf{n}_{\mathbf{x}_t,l}$ at time $t$ in Eq.~\ref{eq:project-point}.

\vspace{-0.5em}
\begin{equation}
\mathbf{n}_{\mathbf{x}_t,l} = \mathbf{n}_{s_l,t} + \frac{|\mathbf{n}_{\mathbf{x}_0,l} - \mathbf{n}_{s_l,0}|}{|\mathbf{n}_{s_l,0} - \mathbf{n}_{e_l,0}|} (\mathbf{n}_{s_l,t} - \mathbf{n}_{e_l,t})
\label{eq:project-point}
\end{equation}

In other words, the projection point $\mathbf{n}_{\mathbf{x}_t,l}$ moves proportionally to the stretching of link $l$. Therefore, the link poses can be defined using the projection point $\mathbf{n}_{\mathbf{x}_t,l}$, decoupling from non-rigid stretching deformations.
Thus, the link pose $P_{\mathcal{L}}(\theta_t, \mathbf{x}_{t})$ is derived via look-at transformation in Eq.~\ref{eq:lookat}. 

\vspace{-0.5em}
\begin{equation}
P_{\mathcal{L}}(\theta_t, \mathbf{x}_t) = \{ \mathscr{A}(\mathbf{n}_{\mathbf{x}_t,l}, \ray(\mathbf{n}_{s_l,t}, \mathbf{n}_{e_l,t})), \forall l \in \mathcal{L} \}
\label{eq:lookat}
\end{equation}

where $\ray(\mathbf{a}, \mathbf{b}) = \frac{\mathbf{b} - \mathbf{a}}{|\mathbf{b} - \mathbf{a}|}$ is the unit direction vector from $\mathbf{a}$ to $\mathbf{b}$. The look-at transformation $\mathscr{A}: \R^3\times S^2 \mapsto \se(3)$ maps the viewing position and direction to a $\se(3)$ pose~\cite{scratchapixel}.   
Note that look-at transformation requires defining an up-direction for each link. We organize the $|\mathcal{L}|$ links into $\frac{|\mathcal{L}|}{2}$ triangles and use their face normals as up-directions. 
Remark that there is a key distinction between our deformable graph and the deformation graph used in Dynamic Fusion~\cite{newcombe2015dynamicfusion}. Dynamic Fusion attaches SE(3) poses to joints only. Our method represents each joint as an $\mathbb{R}^3$ position and derives SE(3) poses at links, which simplifies motion-graph manipulation.

\vspace{0.5em}\noindent\textbf{Graph Connectivity $\mathcal{L}$ Initialization.} The last problem to address is determining the graph connectivity $\mathcal{L}$, which defines the links between joints as integer pairs $l = (s_l, e_l)$, where $s_l, e_l \in [1,|\mathcal{J}_t|]$ denote the indices of the start and end joints. The connectivity $\mathcal{L}$ is treated as a fixed parameter, neither receiving gradients nor changing over time. For kinematic trees, connectivity is derived from domain-specific priors, such as lifting 2D human skeletons to 3D or extracting from robot models. For deformable graphs, we initialize $\mathcal{L}$ by randomly sampling and connecting points from a point cloud, using farthest point sampling~\cite{eldar1997farthest} to ensure uniform coverage of the object’s surface, as illustrated in Fig.~\ref{fig:more-details} and detailed in Appendix ~\ref{sec:optimization}.

\subsection{Weight Painting Function}
\label{sec:m:weipaint}
\vspace{-0.5em}

To estimate the influence of each graph link on a Gaussian, the weight painting function is given as: $\mathcal{W}(\mathbf{x}_{0}, X_{\mathcal{L}, 0}) = \mathrm{softmax}(\{ K(\mathbf{x}_{0}, X_{\mathcal{L}, 0, i}) \mid \forall i \in [1,|\mathcal{L}|] \})$, where $\mathbf{x}_0$ is the initial position of a Gaussian, and $X_{\mathcal{L}, 0, i} \in \mathbb{R}^6$ are the initial positions of start and end joints of the $i$-th link. The kernel function $K: \R^3 \times \R^6 \to \R$ measures the affinity between a Gaussian and a link and is defined as:  

\vspace{-1.5em}
\begin{equation}
K(\mathbf{x}_0, X_{\mathcal{L}, 0, i}) = \exp\left(-\gamma \cdot \text{dist}(\mathbf{x}_0, X_{\mathcal{L}, 0, i})\right)
\label{eq:gk}
\end{equation}

where  $\gamma$ is a learnable parameter that controls the kernel radius, and $\text{dist}(a, b)$ is the distance between point $a$ to line segment $b$. By learning distinct $\gamma$ at each link, the motion graph's influence regions collectively span the object’s surface. 
At $t=0$, 3D Gaussians reconstruct objects' initial geometry. 
The static reconstruction is bonded to the dynamic motion graph through weight painting. 
The painted weights are subsequently used to propagate motions for $t > 0$, for dynamic reconstruction.

\vspace{-1em}
\begin{figure*}[t]
    \centering
    \includegraphics[width=\linewidth]{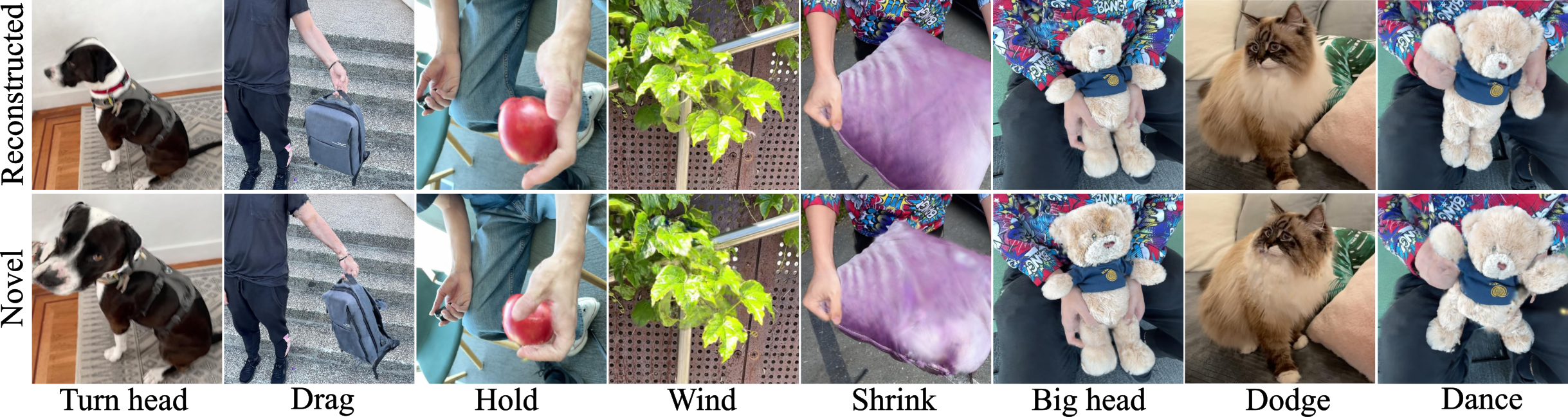}
    \caption{\textbf{Novel Poses from Motion Graph Manipulation.} The first row shows  images of scenes reconstructed from observations. By applying new control actions on the motion graphs and propagating changes to the Gaussians, novel unseen poses are imagined and rendered in the second row.}
    \label{fig:novel-pose}
    \vspace{-0.5em}
\end{figure*}


\begin{figure}[!htp]
    \vspace{-1em}
    \centering
    \captionsetup[subfigure]{justification=centering,labelformat=empty}

    \subfloat[(a) Motion Graph Initialization]{
    \includegraphics[width=0.49\linewidth]{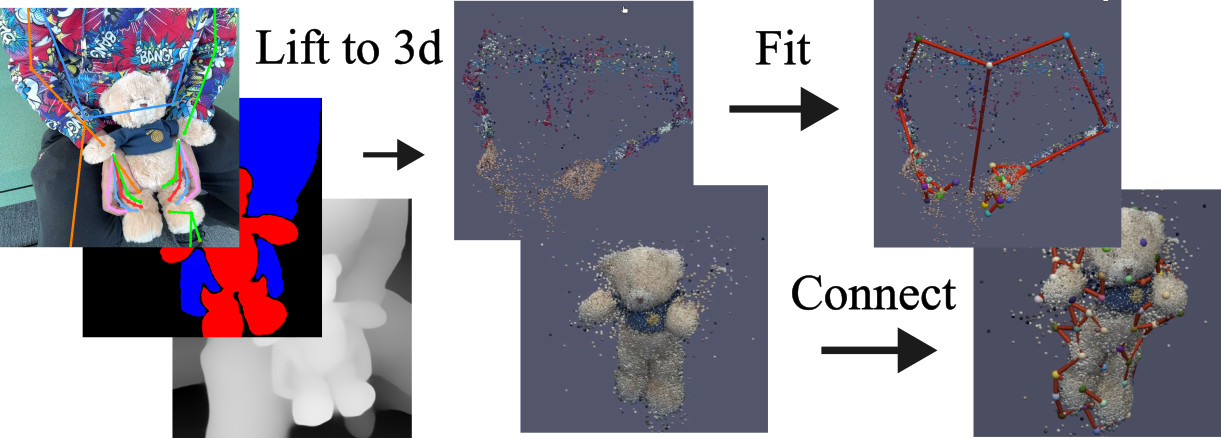}}\hfill
    \subfloat[(b) Instance-level Reconstruction]{
    \includegraphics[width=0.49\linewidth]{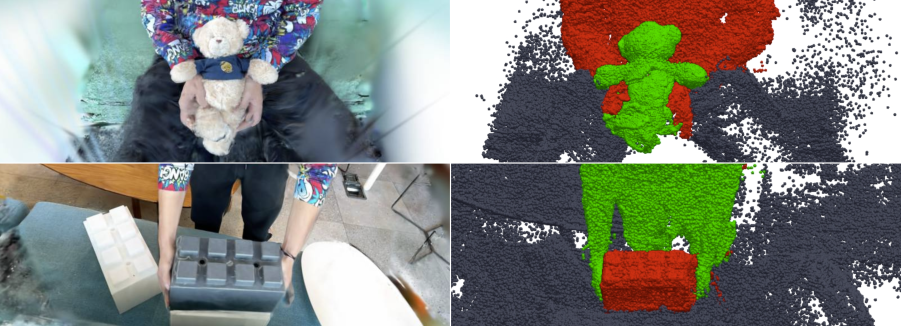}}
    \caption{We initialize motion graphs at the canonical frame ($t=0$) using instance segmentation masks from Grounding SAM2~\cite{ren2024grounded,ravi2024sam2} and 2D human skeletons estimated by SAPIENS~\cite{khirodkar2024sapiens}. Furthermore, our framework enables per-instance reconstruction, where Gaussians are explicitly grouped to maintain accurate instance geometry --- a capability previously unexplored in existing literature of dynamic Gaussian splatting. More optimization details are provided in Appendix~\ref{sec:optimization}.}
    \label{fig:more-details}
    \vspace{-0.5em}
\end{figure}


\vspace{0.5em}
\section{Experiment}



\vspace{-1em}
\subsection{Settings}
\vspace{-0.5em}
\label{sec:e:setting}

\noindent\textbf{Datasets.} We evaluate our method on two real-world datasets: the highly challenging iPhone dataset~\cite{gao2022monocular} from \cite{wang2024shape} and the HyperNeRF~\cite{park2021hypernerf} vrig dataset.
The iPhone dataset consists of 12 scenes, including 5 multi-camera (MV) and 7 single-camera (SV) scenes, each captured at $960\times720$ resolution. The MV scenes are captured using one hand-held moving camera and two fixed cameras at novel angles, where the hand-held video is used for training and the fixed cameras for quantitative evaluation. The SV scenes, captured solely with a hand-held camera, are used for qualitative visualization. 
The HyperNeRF vrig dataset consists of 4 scenes, each captured at $960\times540$ resolution using two cameras mounted on a rig kit with strong camera motions. Models are trained on a subset of frames and evaluated on the remaining frames.

\noindent\textbf{Baselines.} 
On the iPhone dataset, we compare against Shape-of-Motion~\cite{wang2024shape}. For HyperNeRF, we compare against 4DGaussians~\cite{wu20244d}. Shape-of-Motion and 4DGaussians are the state of the art on their respective datasets. We use the  2D track loss proposed in~\cite{wang2024shape}, in addition to L1 RGB loss.

\noindent\textbf{Metrics.} We report peak-signal-to-noise ratio (PSNR), structural similarity index (SSIM)~\cite{wang2004image} and learned perceptual image patch similarity (LPIPS)~\cite{zhang2018unreasonable}. Since traditional metrics such as PSNR and SSIM are sensitive to minor misalignments and often favor blurry images over sharp ones~\cite{park2021hypernerf}, we adopt LPIPS as our primary metric.

\subsection{Results}
\label{sec:e:result}
\vspace{-0.5em}

\begin{figure}[!htp]
\vspace{-1em}
\centering
\begin{minipage}{.65\linewidth}
\captionof{table}{Novel view rendering on the highly challenging iPhone dataset~\cite{wang2024shape}. LPIPS more accurately reflects perceptual quality.}\label{tab:iphone}
\resizebox{0.96\linewidth}{!}{\begin{tabular}{lccc}\toprule 
\textbf{Method} &\textbf{LPIPS $\downarrow$} &\textbf{PSNR $\uparrow$} &\textbf{SSIM $\uparrow$} \\\midrule  
T-NeRF~\cite{gao2022monocular} &0.55 &15.6 &0.55 \\  
HyperNeRF~\cite{park2021hypernerf} &0.51 &15.99 &0.59 \\  
DynIBaR~\cite{li2023dynibar} &0.55 &13.41 &0.48 \\  
Deformable-GS~\cite{Yang2023Deformable3G} &0.66 &11.92 &0.49 \\  
4DGaussians~\cite{wu20244d} &0.56 &13.42 &0.49 \\  
Shape-of-Motion~\cite{wang2024shape} &0.39 &16.67 &\textbf{0.65} \\  
Ours &\textbf{0.37} &\textbf{16.79} &\textbf{0.65} \\  
\bottomrule  
\end{tabular}}
\end{minipage}
\begin{minipage}{.34\linewidth}
    \flushright
    \includegraphics[width=\textwidth]{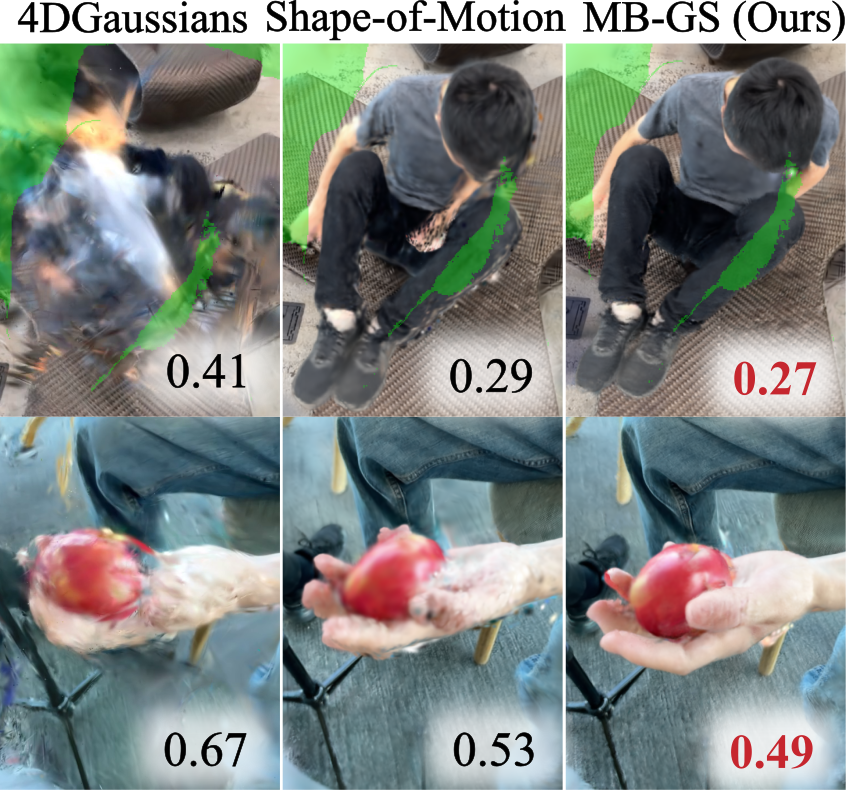}
    \vspace*{-2em}
    \captionsetup{labelfont={scriptsize}}
    \captionof{figure}{\scriptsize 
Novel view rendering on iPhone with best LPIPS in red. More results in Fig.~\ref{fig:quality-compare}.}
    \label{fig:preview-iphone}
\end{minipage}
\vspace{-1em}
\end{figure}

\begin{figure}[!htp]
\centering
\begin{minipage}{0.6\linewidth}
   \captionof{table}{HyperNerf~\cite{park2021hypernerf}. Our method performs competitively, closely matching SoTA in the key LPIPS metric.}\label{tab:hypernerf}
\begin{tabular}{lccc}\toprule
&\textbf{LPIPS $\downarrow$} &\textbf{PSNR $\uparrow$} &\textbf{SSIM $\uparrow$} \\\midrule
4DGaussians~\cite{wu20244d} &0.36 &\textbf{25.19} &\textbf{0.68} \\
Shape-of-Motion~\cite{wang2024shape} &\textbf{0.34} &21.01 &0.54 \\
Ours &0.35 &20.60 &0.54 \\
\bottomrule
\end{tabular} 
\end{minipage}
\begin{minipage}{0.32\linewidth}
         \flushright
    \includegraphics[width=\textwidth]{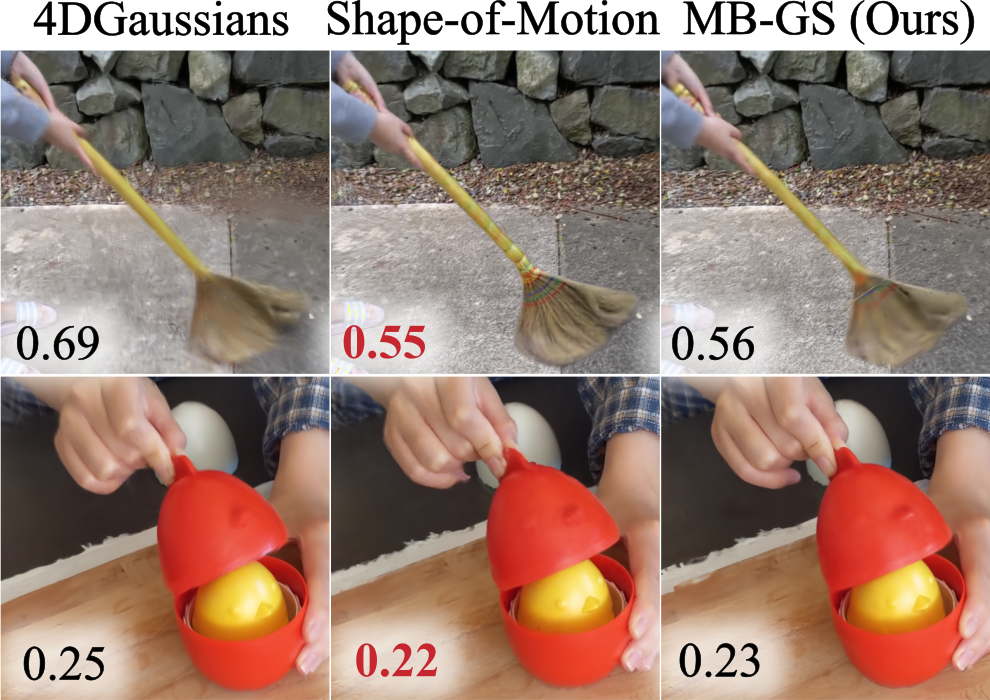}
    \vspace*{-2em}
    \captionsetup{labelfont={scriptsize}}
    \captionof{figure}{\scriptsize 
    {Visualization on HyperNerf with
best LPIPS in red. More results in Fig.~\ref{fig:hypernerf}}}
    \label{fig:preview-hypernerf}
\end{minipage}
\vspace{-1em}
\end{figure}

Tab.~\ref{tab:iphone} presents our quantitative results on the iPhone benchmark, where our method outperforms the state-of-the-art Shape-of-Motion by 0.02 LPIPS. Fig.~\ref{fig:quality-compare} provides a qualitative comparison on the five MV scenes with 4DGaussians~\cite{wu20244d} and Shape-of-Motion~\cite{wang2024shape}, demonstrating that our method renders sharper, more complete, and perceptually higher-quality novel views. While 4DGaussians achieves comparable PSNR and SSIM scores, its rendering quality is notably inferior.
Fig.~\ref{fig:motion-graph-demo} visualizes the learned motion graphs and weight paintings. Training on the iPhone dataset converges within 10 to 30 hours on a 40GB A100 GPU, depending on scene complexity. For the teddy bear scene with 2M Gaussians—using a deformable graph for the bear and a kinematic tree for the human—our method achieves a rendering speed of 18 FPS, and 25 FPS if motion graphs for each frame are pre-computed and cached. On smaller scenes, such as the chicken toy in HyperNeRF with 300K Gaussians, the rendering speed increases to 32 FPS without caching and 46 FPS with caching.

Tab.~\ref{tab:hypernerf} presents our quantitative results on the HyperNeRF vrig benchmark, with qualitative comparison 
in Fig.~\ref{fig:hypernerf}.
While 4DGaussians, the SoTA on HyperNeRF, achieves higher PSNR and SSIM, our method achieves a comparable or better rendering quality and LPIPS on the chicken, 3D printer, and broom scenes (rows 2 to 4 in Fig.~\ref{fig:hypernerf}). 
Notably, 4DGaussians achieves a better score on the 3D printer scene, but our method produces clearer results, accurately rendering the text on the printer motor, unlike the blurry output from 4DGaussians.
However, on the peel-banana scene, our method produces lower-quality results than 4DGaussian. We further discuss our limitations in Sec.~\ref{sec:limitation}.

\subsection{Applications}
\label{sec:applications}
\vspace{-0.5em}

\begin{figure*}[t]
    \vspace{-1.5em}
    \centering
    \includegraphics[width=\linewidth]{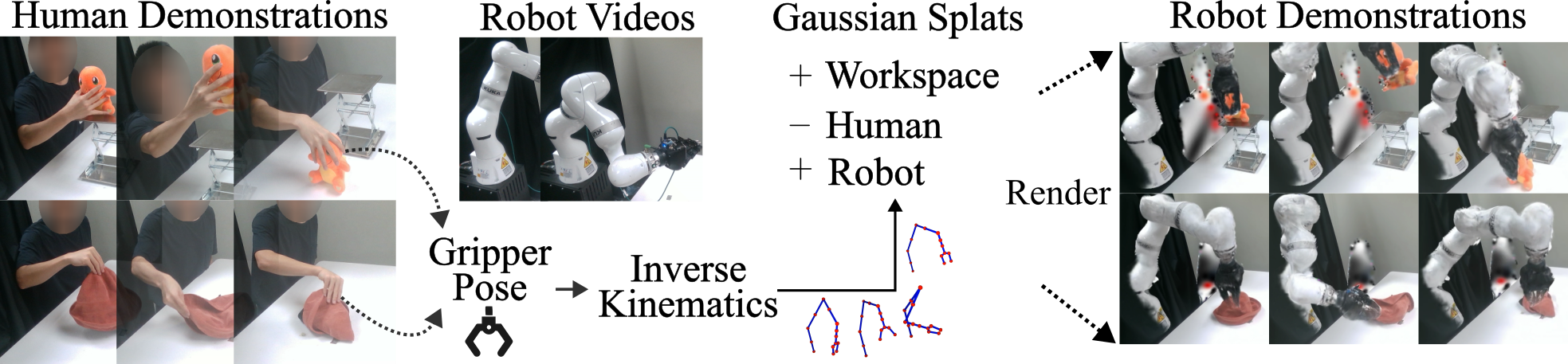}
    \caption{\textbf{Robot Demonstration from Human Videos.} We reconstruct Gaussians from human and robot videos, with robot's pre-defined kinematic chain as motion graph. We then remove human's Gaussians and add robot's to workspace. Next, we animate the robot motion graph using inverse kinematics derived from hand poses. This renders videos of a robot arm mimicking human actions.}
    \label{fig:robot-pipeline}
\end{figure*}

\begin{figure*}[t]
    \vspace{-0.5em}
    \centering
    \includegraphics[width=\linewidth]{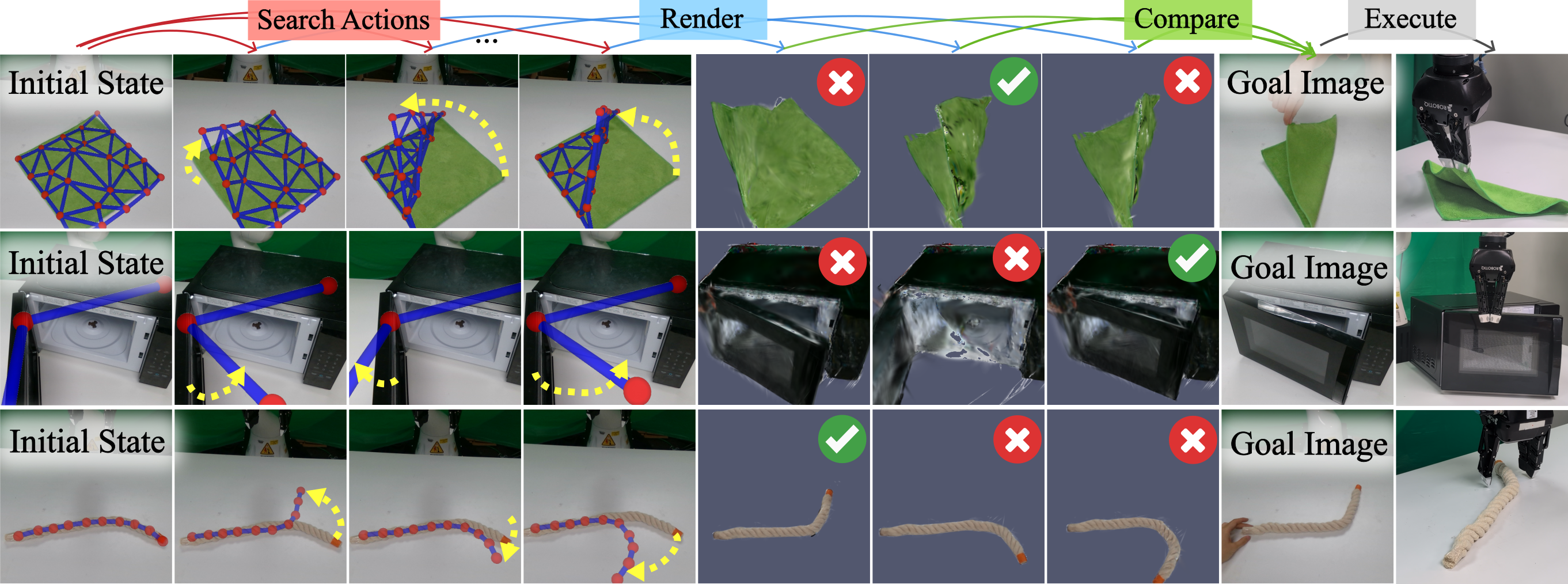}

    \caption{\textbf{Robotic Action Prediction via Visual Planning.} We reconstruct object Gaussians (cloth, microwave, rope) from human videos, simulate potential motion graph trajectories, and render the resulting object images. The optimal trajectory is selected by maximizing PSNR between rendered images and the target goal image. Key joint trajectories (yellow) are mapped to end-effector positions. Videos of our robot experiments can be found at \href{https://anonymous368.github.io/motion-blender}{\scriptsize \texttt{anonymous368.github.io/motion-blender}}.}
    \label{fig:visual-plan}
    \vspace{-1em}
\end{figure*}

\noindent\textbf{Novel Pose Animation.} Fig.~\ref{fig:novel-pose} shows images of objects in novel poses that were not seen in the training videos. 
After reconstructing dynamic scenes, we keep the Gaussians fixed and modify the motion graph starting from a sampled frame.  Specifically, we adjust joint rotation angles in a kinematic tree or manipulate the positions of subsets of joints in a deformable graph. These edits are performed interactively using a custom visual editor that we built on VISER~\cite{nerfstudio-viser}. Once the edits are applied, we propagate the updated graph link poses to the Gaussians and render the scene from the same camera viewpoint as the sampled frame. 
To create short animations, we interpolate joint angle or position changes incrementally and render the scene frame-by-frame.

Fig.~\ref{fig:novel-pose} shows that our method extends beyond reconstructing observed motions, enabling the creation of novel imagined object motions while maintaining high rendering quality. 
Recent advances in controllable image and video editing predominantly rely on large diffusion models~\cite{ceylan2023pix2video, shi2024dragdiffusion, huang2024diffusion}. In contrast, our method introduces a novel approach by operating directly in the 3D space, enabling fast and precise control over object poses without the need for neural networks.

\noindent\textbf{Robot Demonstration Synthesis.} 
Imitation learning enables robots to perform complex tasks by learning from robot demonstrations~\cite{o2024open}. Unlike  human videos, collecting robot demonstrations requires labor-intensive teleoperation~\cite{chi2024universal}. In Fig.~\ref{fig:robot-pipeline}, we demonstrate a prototype of our method for synthesizing robot demonstrations from human videos.
We reconstruct scenes from human and robot videos independently. The human videos feature an operator performing pick-and-place and cloth folding tasks. The robot videos feature a Kuka iiwa robot that moves randomly. We use the robot's kinematic chain as the motion graph. 
We then remove the human Gaussians and add the robot Gaussians to the workspace. Keyframes are manually selected, with hand positions adjusted and mapped to robot gripper poses using our visual editor. Gripper poses between keyframes are interpolated, and converted  into joint angles of the robot's kinematic chain using inverse kinematics. The chain is then used as the robot motion graph for view rendering. 
This generates a video where the robot replaces the human, completing the task as demonstrated. 
While some artifacts and unrealistic interactions still exist, this demonstrates the potential of our method for robotic data synthesis. The videos are obtained with minimal effort, and can be used for learning vision-based policies.


\noindent\textbf{Robot Visual Planning.} Predicting action outcomes is a fundamental human capability that enables complex manipulation skills, such as manipulating objects until a certain goal is reached~\cite{ha2018world}. While recent work has explored video generation models to incorporate this ability for robotic manipulation~\cite{yang2023learning,yang2024video}, these approaches often require complicated pipelines. In Fig.~\ref{fig:visual-plan}, we present a simple visual planning prototype for goal-conditioned manipulation of deformable (cloth and rope) and articulated (microwave door) objects. 
 We first reconstruct object-specific Gaussians from human videos. 
For each test scene, the motion graph is learned by minimizing the L1 rendering loss from our reconstructed Gaussians, while keeping Gaussians and weights fixed. This process typically converges within a minute.
 We then simulate various graph trajectories and render the resulting object images. The optimal trajectory is selected by maximizing PSNR with the provided target goal image. 
In summary, our method learns a dynamic and photorealistic model of the manipulated object {\it on the fly}, using a very small number of frames, and performs planning in simulation with the learned model to select optimal actions. The reward function is simply provided in the form of a goal image. This prototype demonstrates the potential of our method for developing more data-efficient model-based robot learning solutions. More experiment details are provided in Appendix~\ref{apn:robotics-details}.

\vspace{-1em}
\section{Final Remarks}
\vspace{-1em}

We conducted ablation studies analyzing motion graph sizes and regularization strategies. Ablation studies, related work discussions, more visualizations and optimization details  are included in Appendix~\ref{sec:apn} due to space limit. We discuss the limitations and possible future directions in Sec.~\ref{sec:limitation}.



\newpage

\section{Limitations and Future Directions}
\label{sec:limitation}

\begin{figure}[h]
    \centering
    \includegraphics[width=\linewidth]{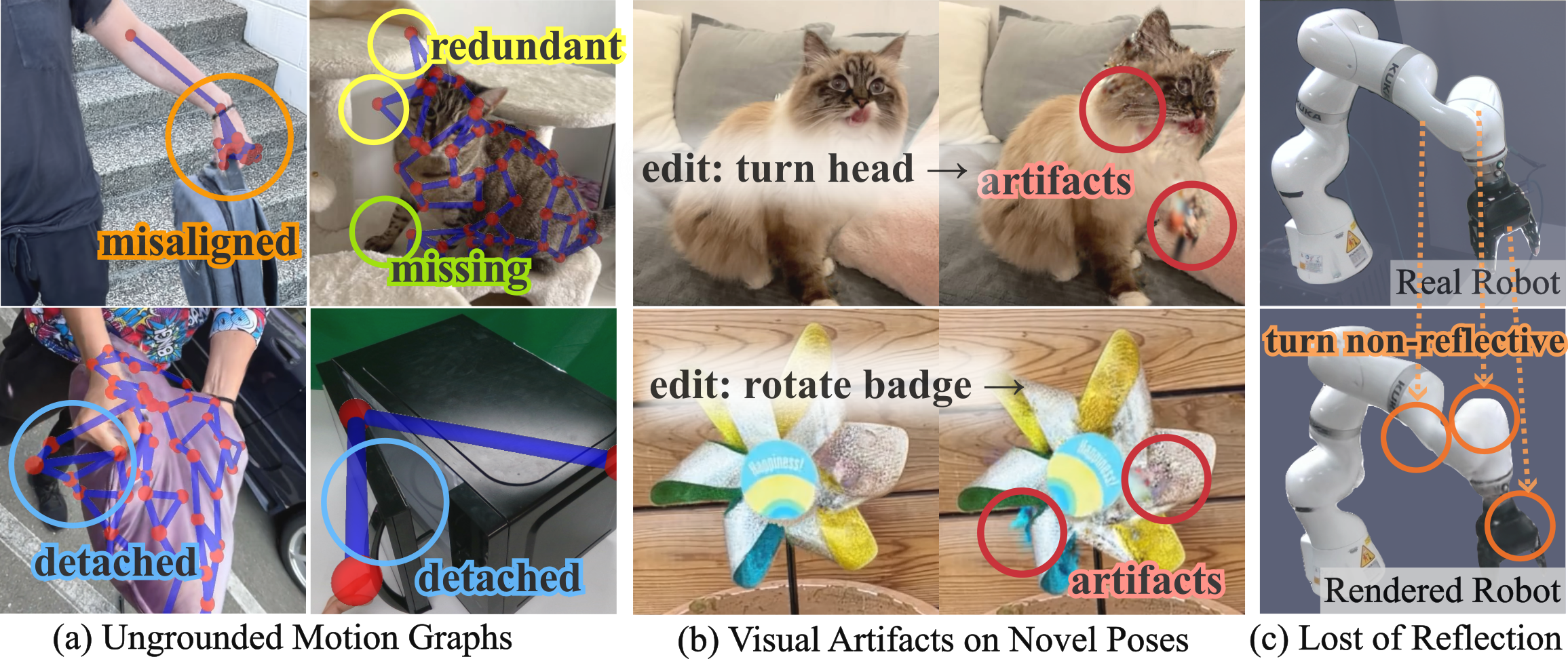}
    \caption{\textbf{Failure Cases.} Visualization of imperfect learned motion graphs (a), failure cases of novel pose editing (b), and failure cases of reconstructing reflective surfaces (c).}
    \label{fig:limitation}
\end{figure}

\noindent\textbf{Visual Artifacts on Novel Poses.}
Fig.~\ref{fig:limitation} (b) demonstrates failure cases in animating novel object poses,  where attempting to turn the cat's head and rotate the windmill's badge introduces non-negligible visual artifacts. Visual artifacts can also be found in Fig.~\ref{fig:robot-pipeline} and Fig.~\ref{fig:visual-plan} of our robot experiments. We believe the key reason is that, unlike meshes, Gaussians lack an explicit \emph{surface representation}. This allows each individual Gaussian to deform arbitrarily when motion graphs are applied. 
This can cause some Gaussians to deviate from object surfaces, particularly for unseen motions. Recent advances like 2D-GS~\cite{huang20242d}, which directly formulates Gaussians on 3D surfaces, offer promising solutions to this limitation.

\noindent\textbf{Ungrounded Motion Graphs.} Fig.~\ref{fig:limitation} (a) demonstrates imperfect learned motion graphs that are not accurately grounded to the object geometry. Top left: the hand contains intricate kinematic structures but only occupies a small region (less than $80\times80$ pixels in a $720\times920$ image), making it difficult to capture fine details. Top right: incomplete motion graph coverage on the cat's right front leg and joints protruding beyond the cat body. Bottom left: the pillow is squeezed but the motion graph fails to sufficiently deform. Bottom right: the kinematic tree does not accurately align with the microwave door.

\begin{figure}[h!]
    \centering
    \includegraphics[width=0.7\linewidth]{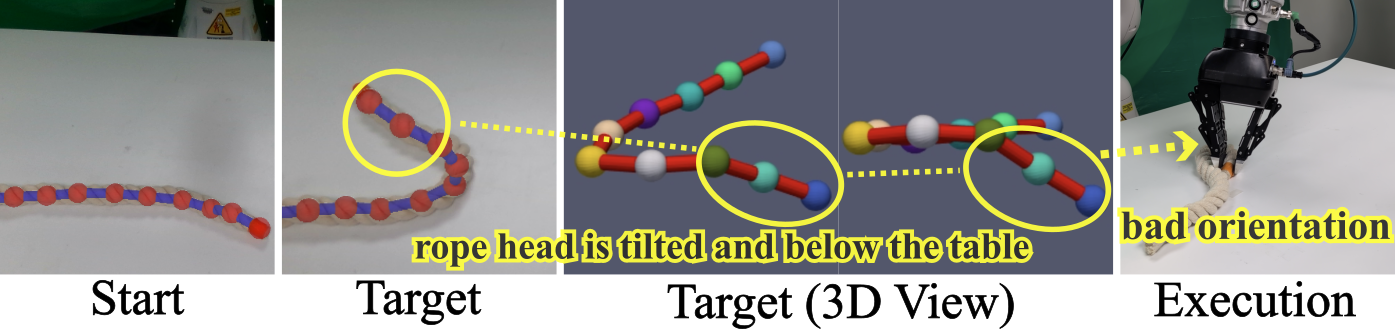}
    \caption{A failure case of incorrect gripper orientation caused by an erroneous motion graph. The head of the rope motion graph is misaligned, tilting downward below the table surface.}
    \label{fig:limitation-robot}
\end{figure}

We believe the key reason is that Gaussian splatting reconstructs scenes purely from visual observations, without incorporating structural or physical priors. 
For instance, Fig.~\ref{fig:limitation-robot} illustrates a failure case in rope manipulation. While the reconstructed motion graph appears accurate in 2D image space, a 3D inspection reveals that the rope's head tilts below the table---an incorrect reconstruction undetectable from visual observation alone. Since the robot's end-effector pose is derived from the motion graph's joints and links, this leads to an erroneous orientation in the planned trajectory.
However, recovering sparse structures from only a few videos is a highly under-constrained optimization problem.  A promising direction is to incorporate semantic or physical priors from foundation models to improve geometric consistency.

\begin{figure}[h]
    \centering
    \includegraphics[width=\linewidth]{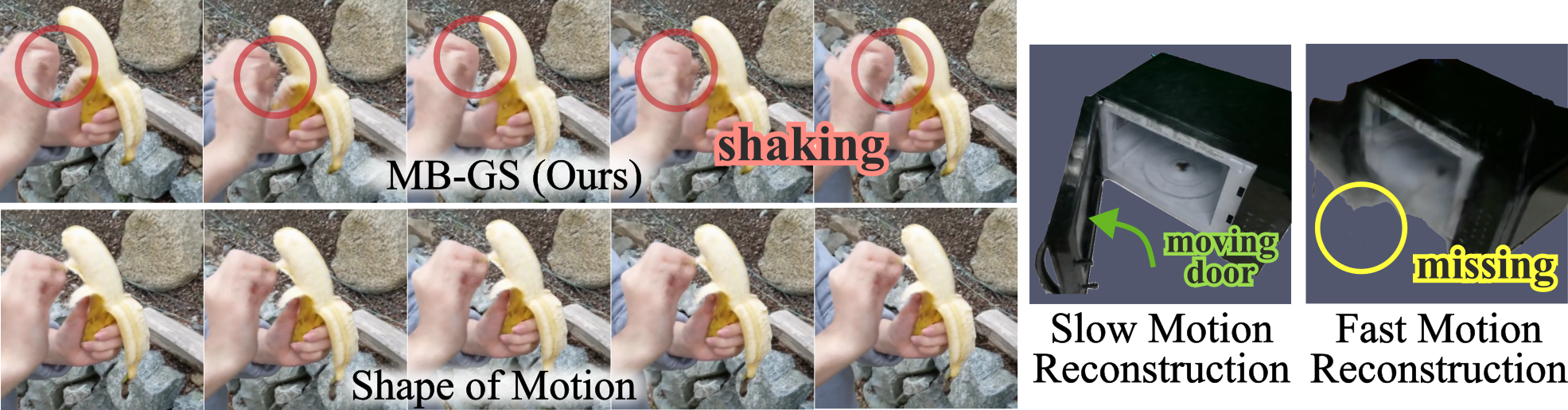}
    \caption{\textbf{Failure Cases on Fast-Moving Objects.} Visualization of shaking artifacts from strong camera motion (left), and reconstruction quality comparison of our MB-GS for the same microwave with doors moving at different speeds (right).}
    \label{fig:limitation-motion}
\end{figure}

\noindent\textbf{Fast Camera and Object Motions.}  
Fig.~\ref{fig:limitation-motion} shows failure cases when reconstructing fast-moving objects. The left figure compares our MB-GS with Shape-of-Motion on adjacent frames from the HyperNerf peel-banana scene, which is captured with strong camera motions and objects (the hand and banana) frequently entering and exiting the view. In this setting, MB-GS produces inconsistent results across frames such as a shaking hand. The right figure compares MB-GS reconstructions of a microwave with doors moving at different speeds. We observe that reconstruction quality degrades significantly when the door moves rapidly.  
Dynamic reconstruction under large motions remains a longstanding challenge~\cite{wu20244d}. Shape-of-Motion attempts to address this by using 2D tracks~\cite{doersch2022tap} as motion guidance. However, 2D tracks are often unreliable, especially under orientation changes, such as when the microwave doors rotate and original tracks on the door surface are lost. Another limitation is that MB-GS learns motion graph parameters independently at each frame, making it less effective at leveraging temporal continuity across frames. A promising direction is to still follow the motion blender framework but introduce neural networks that predict the deformation of motion graphs over time, preserving the manipulability of the representation while improving learning capacity.

\noindent\textbf{Robotic Applications.} In the following, we highlight two promising directions that are specific to the robotic applications of our method MB-GS.

\begin{itemize}
\item \emph{Motion Graphs and Physics Simulation.} Our framework synthesizes robotic demonstrations for tasks like grasping toys and folding clothes (Fig.~\ref{fig:robot-pipeline}), and performs planning in graph space to reach goal configurations (Fig.~\ref{fig:visual-plan}). However, our current motion graph representation lacks physical awareness. This is evident in Fig.~\ref{fig:robot-pipeline} (second row), where the gripper penetrates the cloth in a physically unrealistic manner. Additionally, the graph-space planning is limited to simple shape (2D rectangle) and basic kinematic chain (chain-like graph), whose motion patterns can be reasonably approximated with geometric heuristics.

Our sparse motion graph design naturally extends to physics-based simulation frameworks. For instance, MuJoCo~\cite{todorov2012mujoco} models deformable objects using flex elements (triangles in 2D, tetrahedra in 3D), which aligns closely with our deformable graph representation. To further bridge the gap, it is also possible to, first, incorporate not only link poses but also triangular face poses to guide Gaussian motion, leveraging projection points analogous to Fig.~\ref{fig:motion-graph}; second, support hybrid deformable and kinematic structures through a more systematic engineering integration. These extensions would significantly improve simulation compatibility for both articulated and deformable objects.

In doing so, instead of learning joint positions or rotation angles as in our current approach, it is possible to learn MuJoCo's physical parameters through differentiable simulation: simulating the structure, rendering images, and back-propagating through the physics engine. This would be particularly feasible in controlled environments with clear object segmentation, enabling more powerful digital twins that handle deformable and articulated objects --- a capability lacking in current rigid-body-focused digital twins~\cite{torne2024reconciling}.


\item \emph{Modeling of Light Sources and Reflective Surfaces.} Robotic workspaces typically contain strong light sources, and robot arms often have reflective surfaces. However, as shown in Fig.~\ref{fig:limitation}, our reconstructed robot surfaces appear blurry and lose their reflectivity. This occurs because Gaussian splatting does not account for lighting effects during reconstruction.
While this limitation is acceptable for vanilla Gaussian splatting in static scenes (where lighting effects remain mostly constant), it becomes problematic for dynamic scenes. In such cases, surfaces under varying lighting conditions appear to move constantly, leading to blurry reconstructions.
Furthermore, workspace lighting affects not only robot reconstruction but also object reconstruction, as objects may be reflective or cast different shadows when illuminated from various angles. We find that increased lighting generally results in blurrier reconstructed objects. While this is a fundamental limitation of current Gaussian splatting methods rather than specific to our approach, it highlights the need for reflection-aware Gaussian splatting techniques to improve robotic applications.
\end{itemize}

\newpage


\begin{thebibliography}{69}
\providecommand{\natexlab}[1]{#1}
\providecommand{\url}[1]{\texttt{#1}}
\expandafter\ifx\csname urlstyle\endcsname\relax
  \providecommand{\doi}[1]{doi: #1}\else
  \providecommand{\doi}{doi: \begingroup \urlstyle{rm}\Url}\fi

\bibitem[Wang et~al.(2024)Wang, Ye, Gao, Austin, Li, and Kanazawa]{wang2024shape}
Q.~Wang, V.~Ye, H.~Gao, J.~Austin, Z.~Li, and A.~Kanazawa.
\newblock Shape of motion: 4d reconstruction from a single video.
\newblock \emph{arXiv preprint arXiv:2407.13764}, 2024.

\bibitem[Kerbl et~al.(2023)Kerbl, Kopanas, Leimk{\"u}hler, and Drettakis]{kerbl20233d}
B.~Kerbl, G.~Kopanas, T.~Leimk{\"u}hler, and G.~Drettakis.
\newblock 3d gaussian splatting for real-time radiance field rendering.
\newblock \emph{ACM Trans. Graph.}, 42\penalty0 (4):\penalty0 139--1, 2023.

\bibitem[Wu et~al.(2024)Wu, Yi, Fang, Xie, Zhang, Wei, Liu, Tian, and Wang]{wu20244d}
G.~Wu, T.~Yi, J.~Fang, L.~Xie, X.~Zhang, W.~Wei, W.~Liu, Q.~Tian, and X.~Wang.
\newblock 4d gaussian splatting for real-time dynamic scene rendering.
\newblock In \emph{Proceedings of the IEEE/CVF conference on computer vision and pattern recognition}, pages 20310--20320, 2024.

\bibitem[Li et~al.(2024)Li, Chen, Li, and Xu]{li2024spacetime}
Z.~Li, Z.~Chen, Z.~Li, and Y.~Xu.
\newblock Spacetime gaussian feature splatting for real-time dynamic view synthesis.
\newblock In \emph{Proceedings of the IEEE/CVF Conference on Computer Vision and Pattern Recognition}, pages 8508--8520, 2024.

\bibitem[Yang et~al.(2023)Yang, Yang, Pan, and Zhang]{yang2023real}
Z.~Yang, H.~Yang, Z.~Pan, and L.~Zhang.
\newblock Real-time photorealistic dynamic scene representation and rendering with 4d gaussian splatting.
\newblock \emph{arXiv preprint arXiv:2310.10642}, 2023.

\bibitem[Xie et~al.(2024)Xie, Zong, Qiu, Li, Feng, Yang, and Jiang]{xie2024physgaussian}
T.~Xie, Z.~Zong, Y.~Qiu, X.~Li, Y.~Feng, Y.~Yang, and C.~Jiang.
\newblock Physgaussian: Physics-integrated 3d gaussians for generative dynamics.
\newblock In \emph{Proceedings of the IEEE/CVF Conference on Computer Vision and Pattern Recognition}, pages 4389--4398, 2024.

\bibitem[Huang et~al.(2024)Huang, Sun, Yang, Lyu, Cao, and Qi]{Huang_2024_CVPR}
Y.-H. Huang, Y.-T. Sun, Z.~Yang, X.~Lyu, Y.-P. Cao, and X.~Qi.
\newblock Sc-gs: Sparse-controlled gaussian splatting for editable dynamic scenes.
\newblock In \emph{Proceedings of the IEEE/CVF Conference on Computer Vision and Pattern Recognition (CVPR)}, pages 4220--4230, June 2024.

\bibitem[Yang et~al.(2023)Yang, Gao, Zhou, Jiao, Zhang, and Jin]{Yang2023Deformable3G}
Z.~Yang, X.~Gao, W.~Zhou, S.~Jiao, Y.~Zhang, and X.~Jin.
\newblock Deformable 3d gaussians for high-fidelity monocular dynamic scene reconstruction.
\newblock \emph{2024 IEEE/CVF Conference on Computer Vision and Pattern Recognition (CVPR)}, pages 20331--20341, 2023.
\newblock URL \url{https://api.semanticscholar.org/CorpusID:262466218}.

\bibitem[James and Twigg(2005)]{james2005skinning}
D.~L. James and C.~D. Twigg.
\newblock Skinning mesh animations.
\newblock \emph{ACM Transactions on Graphics (TOG)}, 24\penalty0 (3):\penalty0 399--407, 2005.

\bibitem[Sumner et~al.(2007)Sumner, Schmid, and Pauly]{sumner2007embedded}
R.~W. Sumner, J.~Schmid, and M.~Pauly.
\newblock Embedded deformation for shape manipulation.
\newblock In \emph{ACM siggraph 2007 papers}, pages 80--es. 2007.

\bibitem[Joshi et~al.(2007)Joshi, Meyer, DeRose, Green, and Sanocki]{joshi2007harmonic}
P.~Joshi, M.~Meyer, T.~DeRose, B.~Green, and T.~Sanocki.
\newblock Harmonic coordinates for character articulation.
\newblock \emph{ACM transactions on graphics (TOG)}, 26\penalty0 (3):\penalty0 71--es, 2007.

\bibitem[Nieto and Sus{\'\i}n(2012)]{nieto2012cage}
J.~R. Nieto and A.~Sus{\'\i}n.
\newblock Cage based deformations: a survey.
\newblock In \emph{Deformation Models: Tracking, Animation and Applications}, pages 75--99. Springer, 2012.

\bibitem[Community(2018)]{blender}
B.~O. Community.
\newblock \emph{Blender - a 3D modelling and rendering package}.
\newblock Blender Foundation, Stichting Blender Foundation, Amsterdam, 2018.
\newblock URL \url{http://www.blender.org}.

\bibitem[Kavan et~al.(2007)Kavan, Collins, {\v{Z}}{\'a}ra, and O'Sullivan]{kavan2007skinning}
L.~Kavan, S.~Collins, J.~{\v{Z}}{\'a}ra, and C.~O'Sullivan.
\newblock Skinning with dual quaternions.
\newblock In \emph{Proceedings of the 2007 symposium on Interactive 3D graphics and games}, pages 39--46, 2007.

\bibitem[Gao et~al.(2022)Gao, Li, Tulsiani, Russell, and Kanazawa]{gao2022monocular}
H.~Gao, R.~Li, S.~Tulsiani, B.~Russell, and A.~Kanazawa.
\newblock Monocular dynamic view synthesis: A reality check.
\newblock \emph{Advances in Neural Information Processing Systems}, 35:\penalty0 33768--33780, 2022.

\bibitem[Park et~al.(2021)Park, Sinha, Hedman, Barron, Bouaziz, Goldman, Martin-Brualla, and Seitz]{park2021hypernerf}
K.~Park, U.~Sinha, P.~Hedman, J.~T. Barron, S.~Bouaziz, D.~B. Goldman, R.~Martin-Brualla, and S.~M. Seitz.
\newblock Hypernerf: A higher-dimensional representation for topologically varying neural radiance fields.
\newblock \emph{arXiv preprint arXiv:2106.13228}, 2021.

\bibitem[Walke et~al.(2023)Walke, Black, Zhao, Vuong, Zheng, Hansen-Estruch, He, Myers, Kim, Du, Lee, Fang, Finn, and Levine]{pmlr-v229-walke23a}
H.~R. Walke, K.~Black, T.~Z. Zhao, Q.~Vuong, C.~Zheng, P.~Hansen-Estruch, A.~W. He, V.~Myers, M.~J. Kim, M.~Du, A.~Lee, K.~Fang, C.~Finn, and S.~Levine.
\newblock Bridgedata v2: A dataset for robot learning at scale.
\newblock In J.~Tan, M.~Toussaint, and K.~Darvish, editors, \emph{Proceedings of The 7th Conference on Robot Learning}, volume 229 of \emph{Proceedings of Machine Learning Research}, pages 1723--1736. PMLR, 06--09 Nov 2023.
\newblock URL \url{https://proceedings.mlr.press/v229/walke23a.html}.

\bibitem[Lin et~al.(2024)Lin, Dai, Zhu, and Yao]{lin2024gaussian}
Y.~Lin, Z.~Dai, S.~Zhu, and Y.~Yao.
\newblock Gaussian-flow: 4d reconstruction with dynamic 3d gaussian particle.
\newblock In \emph{Proceedings of the IEEE/CVF Conference on Computer Vision and Pattern Recognition}, pages 21136--21145, 2024.

\bibitem[Kucuk and Bingul(2006)]{kucuk2006robot}
S.~Kucuk and Z.~Bingul.
\newblock \emph{Robot kinematics: Forward and inverse kinematics}.
\newblock INTECH Open Access Publisher London, UK, 2006.

\bibitem[Scratchapixel(2025)]{scratchapixel}
Scratchapixel.
\newblock Framing: The look-at function.
\newblock \url{https://www.scratchapixel.com/lessons/mathematics-physics-for-computer-graphics/lookat-function/framing-lookat-function.html}, 2025.
\newblock Accessed: 2025-02-20.

\bibitem[Newcombe et~al.(2015)Newcombe, Fox, and Seitz]{newcombe2015dynamicfusion}
R.~A. Newcombe, D.~Fox, and S.~M. Seitz.
\newblock Dynamicfusion: Reconstruction and tracking of non-rigid scenes in real-time.
\newblock In \emph{Proceedings of the IEEE conference on computer vision and pattern recognition}, pages 343--352, 2015.

\bibitem[Eldar et~al.(1997)Eldar, Lindenbaum, Porat, and Zeevi]{eldar1997farthest}
Y.~Eldar, M.~Lindenbaum, M.~Porat, and Y.~Y. Zeevi.
\newblock The farthest point strategy for progressive image sampling.
\newblock \emph{IEEE transactions on image processing}, 6\penalty0 (9):\penalty0 1305--1315, 1997.

\bibitem[Ren et~al.(2024)Ren, Liu, Zeng, Lin, Li, Cao, Chen, Huang, Chen, Yan, Zeng, Zhang, Li, Yang, Li, Jiang, and Zhang]{ren2024grounded}
T.~Ren, S.~Liu, A.~Zeng, J.~Lin, K.~Li, H.~Cao, J.~Chen, X.~Huang, Y.~Chen, F.~Yan, Z.~Zeng, H.~Zhang, F.~Li, J.~Yang, H.~Li, Q.~Jiang, and L.~Zhang.
\newblock Grounded sam: Assembling open-world models for diverse visual tasks, 2024.

\bibitem[Ravi et~al.(2024)Ravi, Gabeur, Hu, Hu, Ryali, Ma, Khedr, Rädle, Rolland, Gustafson, Mintun, Pan, Alwala, Carion, Wu, Girshick, Dollár, and Feichtenhofer]{ravi2024sam2}
N.~Ravi, V.~Gabeur, Y.-T. Hu, R.~Hu, C.~Ryali, T.~Ma, H.~Khedr, R.~Rädle, C.~Rolland, L.~Gustafson, E.~Mintun, J.~Pan, K.~V. Alwala, N.~Carion, C.-Y. Wu, R.~Girshick, P.~Dollár, and C.~Feichtenhofer.
\newblock Sam 2: Segment anything in images and videos, 2024.
\newblock URL \url{https://arxiv.org/abs/2408.00714}.

\bibitem[Khirodkar et~al.(2024)Khirodkar, Bagautdinov, Martinez, Zhaoen, James, Selednik, Anderson, and Saito]{khirodkar2024sapiens}
R.~Khirodkar, T.~Bagautdinov, J.~Martinez, S.~Zhaoen, A.~James, P.~Selednik, S.~Anderson, and S.~Saito.
\newblock Sapiens: Foundation for human vision models.
\newblock \emph{arXiv preprint arXiv:2408.12569}, 2024.

\bibitem[Wang et~al.(2004)Wang, Bovik, Sheikh, and Simoncelli]{wang2004image}
Z.~Wang, A.~C. Bovik, H.~R. Sheikh, and E.~P. Simoncelli.
\newblock Image quality assessment: from error visibility to structural similarity.
\newblock \emph{IEEE transactions on image processing}, 13\penalty0 (4):\penalty0 600--612, 2004.

\bibitem[Zhang et~al.(2018)Zhang, Isola, Efros, Shechtman, and Wang]{zhang2018unreasonable}
R.~Zhang, P.~Isola, A.~A. Efros, E.~Shechtman, and O.~Wang.
\newblock The unreasonable effectiveness of deep features as a perceptual metric.
\newblock In \emph{Proceedings of the IEEE conference on computer vision and pattern recognition}, pages 586--595, 2018.

\bibitem[Li et~al.(2023)Li, Wang, Cole, Tucker, and Snavely]{li2023dynibar}
Z.~Li, Q.~Wang, F.~Cole, R.~Tucker, and N.~Snavely.
\newblock Dynibar: Neural dynamic image-based rendering.
\newblock In \emph{Proceedings of the IEEE/CVF Conference on Computer Vision and Pattern Recognition}, pages 4273--4284, 2023.

\bibitem[Group(2023)]{nerfstudio-viser}
N.~S. Group.
\newblock Viser: Web-based 3d visualization + python.
\newblock \url{https://github.com/nerfstudio-project/viser}, 2023.

\bibitem[Ceylan et~al.(2023)Ceylan, Huang, and Mitra]{ceylan2023pix2video}
D.~Ceylan, C.-H.~P. Huang, and N.~J. Mitra.
\newblock Pix2video: Video editing using image diffusion.
\newblock In \emph{Proceedings of the IEEE/CVF International Conference on Computer Vision}, pages 23206--23217, 2023.

\bibitem[Shi et~al.(2024)Shi, Xue, Liew, Pan, Yan, Zhang, Tan, and Bai]{shi2024dragdiffusion}
Y.~Shi, C.~Xue, J.~H. Liew, J.~Pan, H.~Yan, W.~Zhang, V.~Y. Tan, and S.~Bai.
\newblock Dragdiffusion: Harnessing diffusion models for interactive point-based image editing.
\newblock In \emph{Proceedings of the IEEE/CVF Conference on Computer Vision and Pattern Recognition}, pages 8839--8849, 2024.

\bibitem[Huang et~al.(2024)Huang, Huang, Liu, Yan, Lv, Liu, Xiong, Zhang, Chen, and Cao]{huang2024diffusion}
Y.~Huang, J.~Huang, Y.~Liu, M.~Yan, J.~Lv, J.~Liu, W.~Xiong, H.~Zhang, S.~Chen, and L.~Cao.
\newblock Diffusion model-based image editing: A survey.
\newblock \emph{arXiv preprint arXiv:2402.17525}, 2024.

\bibitem[O’Neill et~al.(2024)O’Neill, Rehman, Maddukuri, Gupta, Padalkar, Lee, Pooley, Gupta, Mandlekar, Jain, et~al.]{o2024open}
A.~O’Neill, A.~Rehman, A.~Maddukuri, A.~Gupta, A.~Padalkar, A.~Lee, A.~Pooley, A.~Gupta, A.~Mandlekar, A.~Jain, et~al.
\newblock Open x-embodiment: Robotic learning datasets and rt-x models: Open x-embodiment collaboration 0.
\newblock In \emph{2024 IEEE International Conference on Robotics and Automation (ICRA)}, pages 6892--6903. IEEE, 2024.

\bibitem[Chi et~al.(2024)Chi, Xu, Pan, Cousineau, Burchfiel, Feng, Tedrake, and Song]{chi2024universal}
C.~Chi, Z.~Xu, C.~Pan, E.~Cousineau, B.~Burchfiel, S.~Feng, R.~Tedrake, and S.~Song.
\newblock Universal manipulation interface: In-the-wild robot teaching without in-the-wild robots.
\newblock \emph{arXiv preprint arXiv:2402.10329}, 2024.

\bibitem[Ha and Schmidhuber(2018)]{ha2018world}
D.~Ha and J.~Schmidhuber.
\newblock World models.
\newblock \emph{arXiv preprint arXiv:1803.10122}, 2018.

\bibitem[Yang et~al.(2023)Yang, Du, Ghasemipour, Tompson, Schuurmans, and Abbeel]{yang2023learning}
M.~Yang, Y.~Du, K.~Ghasemipour, J.~Tompson, D.~Schuurmans, and P.~Abbeel.
\newblock Learning interactive real-world simulators.
\newblock \emph{arXiv preprint arXiv:2310.06114}, 1\penalty0 (2):\penalty0 6, 2023.

\bibitem[Yang et~al.(2024)Yang, Walker, Parker-Holder, Du, Bruce, Barreto, Abbeel, and Schuurmans]{yang2024video}
S.~Yang, J.~Walker, J.~Parker-Holder, Y.~Du, J.~Bruce, A.~Barreto, P.~Abbeel, and D.~Schuurmans.
\newblock Video as the new language for real-world decision making.
\newblock \emph{arXiv preprint arXiv:2402.17139}, 2024.

\bibitem[Huang et~al.(2024)Huang, Yu, Chen, Geiger, and Gao]{huang20242d}
B.~Huang, Z.~Yu, A.~Chen, A.~Geiger, and S.~Gao.
\newblock 2d gaussian splatting for geometrically accurate radiance fields.
\newblock In \emph{ACM SIGGRAPH 2024 conference papers}, pages 1--11, 2024.

\bibitem[Doersch et~al.(2022)Doersch, Gupta, Markeeva, Recasens, Smaira, Aytar, Carreira, Zisserman, and Yang]{doersch2022tap}
C.~Doersch, A.~Gupta, L.~Markeeva, A.~Recasens, L.~Smaira, Y.~Aytar, J.~Carreira, A.~Zisserman, and Y.~Yang.
\newblock {TAP}-vid: A benchmark for tracking any point in a video.
\newblock \emph{Advances in Neural Information Processing Systems}, 35:\penalty0 13610--13626, 2022.

\bibitem[Todorov et~al.(2012)Todorov, Erez, and Tassa]{todorov2012mujoco}
E.~Todorov, T.~Erez, and Y.~Tassa.
\newblock Mujoco: A physics engine for model-based control.
\newblock In \emph{2012 IEEE/RSJ international conference on intelligent robots and systems}, pages 5026--5033. IEEE, 2012.

\bibitem[Torne et~al.(2024)Torne, Simeonov, Li, Chan, Chen, Gupta, and Agrawal]{torne2024reconciling}
M.~Torne, A.~Simeonov, Z.~Li, A.~Chan, T.~Chen, A.~Gupta, and P.~Agrawal.
\newblock Reconciling reality through simulation: A real-to-sim-to-real approach for robust manipulation.
\newblock \emph{arXiv preprint arXiv:2403.03949}, 2024.

\bibitem[Samavati and Soryani(2023)]{samavati2023deep}
T.~Samavati and M.~Soryani.
\newblock Deep learning-based 3d reconstruction: a survey.
\newblock \emph{Artificial Intelligence Review}, 56\penalty0 (9):\penalty0 9175--9219, 2023.

\bibitem[Zhou et~al.(2024)Zhou, Wu, Zuo, Chen, and Hu]{zhou2024comprehensive}
L.~Zhou, G.~Wu, Y.~Zuo, X.~Chen, and H.~Hu.
\newblock A comprehensive review of vision-based 3d reconstruction methods.
\newblock \emph{Sensors}, 24\penalty0 (7):\penalty0 2314, 2024.

\bibitem[Chang and Boularias(2022)]{Chang2022ScenelevelTA}
H.~Chang and A.~Boularias.
\newblock Scene-level tracking and reconstruction without object priors.
\newblock \emph{2022 IEEE/RSJ International Conference on Intelligent Robots and Systems (IROS)}, pages 3785--3792, 2022.
\newblock URL \url{https://api.semanticscholar.org/CorpusID:251308689}.

\bibitem[Chang et~al.(2023)Chang, Ramesh, Geng, Gan, and Boularias]{Chang2023MonoSTARMS}
H.~Chang, D.~M. Ramesh, S.~Geng, Y.~Gan, and A.~Boularias.
\newblock Mono-star: Mono-camera scene-level tracking and reconstruction.
\newblock \emph{2023 IEEE International Conference on Robotics and Automation (ICRA)}, pages 820--826, 2023.
\newblock URL \url{https://api.semanticscholar.org/CorpusID:256416010}.

\bibitem[Gao and Tedrake(2019)]{gao2019surfelwarp}
W.~Gao and R.~Tedrake.
\newblock Surfelwarp: Efficient non-volumetric single view dynamic reconstruction, 2019.
\newblock URL \url{https://arxiv.org/abs/1904.13073}.

\bibitem[Pfister et~al.(2000)Pfister, Zwicker, Van~Baar, and Gross]{pfister2000surfels}
H.~Pfister, M.~Zwicker, J.~Van~Baar, and M.~Gross.
\newblock Surfels: Surface elements as rendering primitives.
\newblock In \emph{Proceedings of the 27th annual conference on Computer graphics and interactive techniques}, pages 335--342, 2000.

\bibitem[Mildenhall et~al.(2021)Mildenhall, Srinivasan, Tancik, Barron, Ramamoorthi, and Ng]{mildenhall2021nerf}
B.~Mildenhall, P.~P. Srinivasan, M.~Tancik, J.~T. Barron, R.~Ramamoorthi, and R.~Ng.
\newblock Nerf: Representing scenes as neural radiance fields for view synthesis.
\newblock \emph{Communications of the ACM}, 65\penalty0 (1):\penalty0 99--106, 2021.

\bibitem[Park et~al.(2021)Park, Sinha, Barron, Bouaziz, Goldman, Seitz, and Martin-Brualla]{Park_2021_ICCV}
K.~Park, U.~Sinha, J.~T. Barron, S.~Bouaziz, D.~B. Goldman, S.~M. Seitz, and R.~Martin-Brualla.
\newblock Nerfies: Deformable neural radiance fields.
\newblock In \emph{Proceedings of the IEEE/CVF International Conference on Computer Vision (ICCV)}, pages 5865--5874, October 2021.

\bibitem[Pumarola et~al.(2021)Pumarola, Corona, Pons-Moll, and Moreno-Noguer]{Pumarola_2021_CVPR}
A.~Pumarola, E.~Corona, G.~Pons-Moll, and F.~Moreno-Noguer.
\newblock D-nerf: Neural radiance fields for dynamic scenes.
\newblock In \emph{Proceedings of the IEEE/CVF Conference on Computer Vision and Pattern Recognition (CVPR)}, pages 10318--10327, June 2021.

\bibitem[Cao and Johnson(2023)]{cao2023hexplane}
A.~Cao and J.~Johnson.
\newblock Hexplane: A fast representation for dynamic scenes.
\newblock In \emph{Proceedings of the IEEE/CVF Conference on Computer Vision and Pattern Recognition}, pages 130--141, 2023.

\bibitem[Fridovich-Keil et~al.(2023)Fridovich-Keil, Meanti, Warburg, Recht, and Kanazawa]{Keil_2023_CVPR}
S.~Fridovich-Keil, G.~Meanti, F.~R. Warburg, B.~Recht, and A.~Kanazawa.
\newblock K-planes: Explicit radiance fields in space, time, and appearance.
\newblock In \emph{Proceedings of the IEEE/CVF Conference on Computer Vision and Pattern Recognition (CVPR)}, pages 12479--12488, June 2023.

\bibitem[Li et~al.(2023)Li, Sun, Zheng, Wang, Zhang, and Liu]{li2023animatable}
Z.~Li, Y.~Sun, Z.~Zheng, L.~Wang, S.~Zhang, and Y.~Liu.
\newblock Animatable and relightable gaussians for high-fidelity human avatar modeling.
\newblock \emph{arXiv preprint arXiv:2311.16096}, 2023.

\bibitem[Yuan et~al.(2024)Yuan, Li, Huang, De~Mello, Nagano, Kautz, and Iqbal]{yuan2024gavatar}
Y.~Yuan, X.~Li, Y.~Huang, S.~De~Mello, K.~Nagano, J.~Kautz, and U.~Iqbal.
\newblock Gavatar: Animatable 3d gaussian avatars with implicit mesh learning.
\newblock In \emph{Proceedings of the IEEE/CVF Conference on Computer Vision and Pattern Recognition}, pages 896--905, 2024.

\bibitem[Qian et~al.(2024)Qian, Wang, Mihajlovic, Geiger, and Tang]{qian20243dgs}
Z.~Qian, S.~Wang, M.~Mihajlovic, A.~Geiger, and S.~Tang.
\newblock 3dgs-avatar: Animatable avatars via deformable 3d gaussian splatting.
\newblock In \emph{Proceedings of the IEEE/CVF conference on computer vision and pattern recognition}, pages 5020--5030, 2024.

\bibitem[Loper et~al.(2023)Loper, Mahmood, Romero, Pons-Moll, and Black]{loper2023smpl}
M.~Loper, N.~Mahmood, J.~Romero, G.~Pons-Moll, and M.~J. Black.
\newblock Smpl: A skinned multi-person linear model.
\newblock In \emph{Seminal Graphics Papers: Pushing the Boundaries, Volume 2}, pages 851--866. 2023.

\bibitem[Lei et~al.(2024)Lei, Weng, Harley, Guibas, and Daniilidis]{lei2024moscadynamicgaussianfusion}
J.~Lei, Y.~Weng, A.~Harley, L.~Guibas, and K.~Daniilidis.
\newblock Mosca: Dynamic gaussian fusion from casual videos via 4d motion scaffolds, 2024.
\newblock URL \url{https://arxiv.org/abs/2405.17421}.

\bibitem[Kuai et~al.(2023)Kuai, Karthikeyan, Kant, Mirzaei, and Gilitschenski]{kuai2023camm}
T.~Kuai, A.~Karthikeyan, Y.~Kant, A.~Mirzaei, and I.~Gilitschenski.
\newblock Camm: Building category-agnostic and animatable 3d models from monocular videos.
\newblock In \emph{Proceedings of the IEEE/CVF Conference on Computer Vision and Pattern Recognition}, pages 6587--6597, 2023.

\bibitem[Zheng et~al.(2024)Zheng, Chen, Zheng, Gu, Yang, Jin, Li, Zhong, Wang, Liu, et~al.]{zheng2024gaussiangrasper}
Y.~Zheng, X.~Chen, Y.~Zheng, S.~Gu, R.~Yang, B.~Jin, P.~Li, C.~Zhong, Z.~Wang, L.~Liu, et~al.
\newblock Gaussiangrasper: 3d language gaussian splatting for open-vocabulary robotic grasping.
\newblock \emph{IEEE Robotics and Automation Letters}, 2024.

\bibitem[Shorinwa et~al.(2024)Shorinwa, Tucker, Smith, Swann, Chen, Firoozi, Kennedy~III, and Schwager]{shorinwa2024splat}
O.~Shorinwa, J.~Tucker, A.~Smith, A.~Swann, T.~Chen, R.~Firoozi, M.~Kennedy~III, and M.~Schwager.
\newblock Splat-mover: Multi-stage, open-vocabulary robotic manipulation via editable gaussian splatting.
\newblock \emph{arXiv preprint arXiv:2405.04378}, 2024.

\bibitem[Lu et~al.(2024)Lu, Zhang, Wang, Liu, Lu, and Tang]{lu2024manigaussian}
G.~Lu, S.~Zhang, Z.~Wang, C.~Liu, J.~Lu, and Y.~Tang.
\newblock Manigaussian: Dynamic gaussian splatting for multi-task robotic manipulation.
\newblock In \emph{European Conference on Computer Vision}, pages 349--366. Springer, 2024.

\bibitem[Abou-Chakra et~al.(2024)Abou-Chakra, Rana, Dayoub, and Suenderhauf]{abou-chakra2024physically}
J.~Abou-Chakra, K.~Rana, F.~Dayoub, and N.~Suenderhauf.
\newblock Physically embodied gaussian splatting: A realtime correctable world model for robotics.
\newblock In \emph{8th Annual Conference on Robot Learning}, 2024.
\newblock URL \url{https://openreview.net/forum?id=AEq0onGrN2}.

\bibitem[Jiang et~al.(2025)Jiang, Hsu, Zhang, Yu, Wang, and Li]{jiang2025phystwin}
H.~Jiang, H.-Y. Hsu, K.~Zhang, H.-N. Yu, S.~Wang, and Y.~Li.
\newblock Phystwin: Physics-informed reconstruction and simulation of deformable objects from videos.
\newblock \emph{arXiv preprint arXiv:2503.17973}, 2025.

\bibitem[Liu et~al.(2023)Liu, Zeng, Ren, Li, Zhang, Yang, Li, Yang, Su, Zhu, et~al.]{liu2023grounding}
S.~Liu, Z.~Zeng, T.~Ren, F.~Li, H.~Zhang, J.~Yang, C.~Li, J.~Yang, H.~Su, J.~Zhu, et~al.
\newblock Grounding dino: Marrying dino with grounded pre-training for open-set object detection.
\newblock \emph{arXiv preprint arXiv:2303.05499}, 2023.

\bibitem[Kirillov et~al.(2023)Kirillov, Mintun, Ravi, Mao, Rolland, Gustafson, Xiao, Whitehead, Berg, Lo, Doll{\'a}r, and Girshick]{kirillov2023segany}
A.~Kirillov, E.~Mintun, N.~Ravi, H.~Mao, C.~Rolland, L.~Gustafson, T.~Xiao, S.~Whitehead, A.~C. Berg, W.-Y. Lo, P.~Doll{\'a}r, and R.~Girshick.
\newblock Segment anything.
\newblock \emph{arXiv:2304.02643}, 2023.

\bibitem[Ren et~al.(2024)Ren, Jiang, Liu, Zeng, Liu, Gao, Huang, Ma, Jiang, Chen, Xiong, Zhang, Li, Tang, Yu, and Zhang]{ren2024grounding}
T.~Ren, Q.~Jiang, S.~Liu, Z.~Zeng, W.~Liu, H.~Gao, H.~Huang, Z.~Ma, X.~Jiang, Y.~Chen, Y.~Xiong, H.~Zhang, F.~Li, P.~Tang, K.~Yu, and L.~Zhang.
\newblock Grounding dino 1.5: Advance the "edge" of open-set object detection, 2024.

\bibitem[Jiang et~al.(2024)Jiang, Li, Zeng, Ren, Liu, and Zhang]{jiang2024trex2}
Q.~Jiang, F.~Li, Z.~Zeng, T.~Ren, S.~Liu, and L.~Zhang.
\newblock T-rex2: Towards generic object detection via text-visual prompt synergy, 2024.

\bibitem[Renaud et~al.()Renaud, Smeets, and Corbijn~van Willenswaard]{nanomesh}
N.~Renaud, S.~Smeets, and L.~J. Corbijn~van Willenswaard.
\newblock {nanomesh}.
\newblock URL \url{https://github.com/hpgem/nanomesh}.

\bibitem[Karaev et~al.(2024)Karaev, Makarov, Wang, Neverova, Vedaldi, and Rupprecht]{karaev2024cotracker3}
N.~Karaev, I.~Makarov, J.~Wang, N.~Neverova, A.~Vedaldi, and C.~Rupprecht.
\newblock Cotracker3: Simpler and better point tracking by pseudo-labelling real videos.
\newblock \emph{arXiv preprint arXiv:2410.11831}, 2024.

\end{thebibliography}

\appendix

\newpage
\section{Appendix}
\label{sec:apn}
\renewcommand{\thefigure}{A\arabic{figure}}
\setcounter{figure}{0}
\renewcommand{\thetable}{A\arabic{table}}
\setcounter{table}{0}

\subsection{Related Works}

\noindent\textbf{Dynamic Reconstruction.} Dynamic reconstruction aims at recovering the geometry, appearance and motion of dynamic scenes from visual data~\cite{samavati2023deep, zhou2024comprehensive,Chang2022ScenelevelTA,Chang2023MonoSTARMS}. Early approaches, such as DynamicFusion~\cite{newcombe2015dynamicfusion} and SurfelWarp~\cite{gao2019surfelwarp},  rely on 3D representations like the truncated signed distance function (TSDF) or surfel~\cite{pfister2000surfels}, coupled    with explicit deformation graphs to model motion. Recent advances have shifted toward using NeRF~\cite{mildenhall2021nerf, park2021hypernerf, Park_2021_ICCV, Pumarola_2021_CVPR, cao2023hexplane, Keil_2023_CVPR} and 3D Gaussian~\cite{kerbl20233d, wu20244d, li2024spacetime, yang2023real, xie2024physgaussian} as 3D representations. Gaussian splatting-based methods, in particular, have gained rising attention due to their ability of high-quality reconstruction and real-time rendering. For example, 4DGaussians~\cite{wu20244d} and Deformable-GS~\cite{Yang2023Deformable3G} encode motion implicitly using deformation networks. Shape-of-Motion~\cite{wang2024shape}, GaussianFlow~\cite{lin2024gaussian} and STG~\cite{li2024spacetime}  employ shallower models, such as polynomials, which require dense per-Gaussian motion parameters.  In contrast to these approaches, our method introduces a sparse and explicit graph-based motion representation for Gaussian splatting, which enables the reconstruction of realistic and complex scenes with greater interpretability and direct and explicit control of the animation.

\noindent\textbf{Explicit Motion Representations.} Explicit motion representations offer intuitive visualization and straightforward motion manipulation. 
Gaussian-based avatar modeling methods~\cite{li2023animatable, yuan2024gavatar, qian20243dgs} rely on parametric human templates, such as SMPL~\cite{loper2023smpl}, but are inherently limited to human subjects and cannot generalize to arbitrary dynamic objects.
SC-GS~\cite{Huang_2024_CVPR} introduces sparse control points, which can be dragged to create new motions. But SC-GS is restricted to simple, synthetic scenes like those in D-NeRF~\cite{Pumarola_2021_CVPR}. 
MoSca~\cite{lei2024moscadynamicgaussianfusion} employs a scaffold structure to represent motions and deformations.
CAMM~\cite{kuai2023camm} employs kinematic chains for motion representation but is limited to meshes instead of 3D Gaussians and requires occlusion-free videos with clean backgrounds. 
On the other hand, classical animation techniques~\cite{james2005skinning} utilize explicit motion representations such as deformation graphs~\cite{sumner2007embedded}, harmonic coordinates~\cite{joshi2007harmonic}, and cage~\cite{nieto2012cage}. 
However, these methods only focus on  applying manually designed motions to mesh surfaces and are not designed for reconstructing motion or geometry from unstructured video data. In contrast, our method introduces a graph-based motion representation for Gaussian splatting, where motion is propagated from sparse graph links to individual Gaussians.
3D Gaussians and motion graphs are jointly  optimized from videos of realistic and complex scenes.

\begin{figure}[h!]
    \centering
    \includegraphics[width=\linewidth]{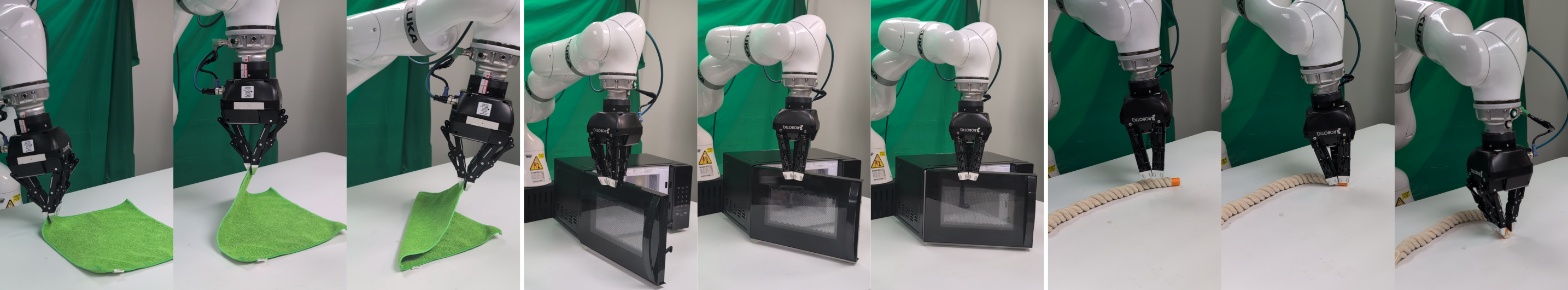}
    \caption{Demonstration of our KUKA robot arm performing cloth folding (left), microwave door adjusting (middle) and rope bending (right), based on planning with our reconstructed motion graphs, without requiring any robot teleoperation data.}
    \label{fig:kuka-robot}
\end{figure}

\noindent\textbf{Robotics Applications of Dynamic Gaussian Splatting.} Recent work has increasingly adopted Gaussian splatting for robotic perception and manipulation. Popular approaches like GaussianGrasper~\cite{zheng2024gaussiangrasper} and SplatMover~\cite{shorinwa2024splat} focus on static reconstructions of rigid objects for digital twin applications. For dynamic scenes, ManiGaussian~\cite{lu2024manigaussian} introduces a Gaussian-based world model for manipulation tasks, while EmbodiedGaussians~\cite{abou-chakra2024physically} incorporates particle physics for interaction modeling. PhysTwin~\cite{jiang2025phystwin} further advances deformable object reconstruction through mass-spring physics. Despite their impressive results, our work differs in two key aspects:
(1) We propose a sparse motion graph representation, while all of these existing work still use dense per-Gaussian motions.
(2) Our framework generalizes beyond constrained robotic workspace settings to complex arbitrary scene. We validate our approach on challenging vision benchmarks including iPhone~\cite{gao2022monocular} and HyperNeRF~\cite{park2021hypernerf}, demonstrating broader applicability.

\subsection{Optimization Details}
\label{sec:optimization}


To reconstruct dynamic scenes from videos, we follow common practices~\cite{wang2024shape, newcombe2015dynamicfusion} and select the frame where objects occupy the largest image area or the frame with most valid 2D tracks as the canonical frame instead of using the first frame. For simplicity, we still use $t=0$ to denote canonical frame. 
We extract instance segmentation masks using Grounding SAM2~\cite{liu2023grounding,kirillov2023segany,ren2024grounding,jiang2024trex2,ren2024grounded,ravi2024sam2} and 2D human skeletons using SAPIENS~\cite{khirodkar2024sapiens}. These annotations are used to initialize the motion graph for each instance at the canonical frame. 
Specifically, the motion graph is parameterized by $(\theta_t)_{t=0}^{T-1}$ (time-varying parameters) and $\phi$ (time-independent parameters). 
We create the graph for each instance at the canonical frame with $(\theta_0, \phi)$ and propagate the parameters across time by setting $\theta_t = \theta_0, \forall t \in [0, T-1]$. We illustrate the motion graph initialization procedure for the canonical frame in Fig.~\ref{fig:more-details} (a).
The 3D Gaussians are initialized for every instance and background using procedure from prior work~\cite{wang2024shape}. 
At each iteration, we transform the 3D Gaussians from the canonical frame $t=0$ to a target frame $t$ using Eq.~\ref{eq:framework}, and minimize the rendering loss against the corresponding video frame.

\textbf{Motion Graph Initialization.} At $t=0$, we lift 2D skeletons and instance masks to 3D point clouds using depth images. For kinematic structures such as the human body, we leverage prior knowledge of joint connections to initialize a kinematic tree with standard joint rotations and link lengths. This tree is then fitted to the point cloud skeleton by minimizing the average point-to-link distance via gradient descent. For deformable graphs, we sample points from the instance point cloud using farthest point sampling~\cite{eldar1997farthest} and connect adjacent points to form the graph.

\noindent\textbf{Instance Level Reconstruction.} To reconstruct each instance, we construct a binary matrix $M \in \mathbb{R}^{|\mathcal{G}|\times I}$, where $|\mathcal{G}|$ denotes the number of Gaussians and $I$ denotes the number of instances. $M$ is defined in Eq.~\ref{eq:instance}.

\begin{equation}
M_{ij} = \begin{cases}
    1 & \text{$i$-th Gaussian belongs to instance $j$} \\
   0 & \text{otherwise} 
\end{cases}
\label{eq:instance}
\end{equation}


The $i$-th row of $M$ represents the one-hot encoding of the instance index for each Gaussian. We then splat $M$ into a 2D instance mask of dimension $\mathbb{R}^{H\times W\times I}$ and minimize the L1 distance between this rendered mask and the instance mask predicted by Grounding SAM2. In doing so, 3D positions of dynamic Gaussians are grouped to always faithfully represent the geometry of each instance, as illustrated in Fig.~\ref{fig:more-details} (b). To the best of our knowledge, our work is the first to study simultaneous per-instance reconstruction in dynamic Gaussian splatting.

\noindent\textbf{Canonical Frame Regularization.} To ensure the learned motion graph remains closely aligned with the object geometry, we minimizes $\sum_{i=1}^J \|\mathbf{n}_{i,0} - \hat{\mathbf{n}}_{i,0}\|$ during training, where $\mathbf{n}_{i,0}$ denotes the position of the current $i$-th joint at the canonical frame, and $\hat{\mathbf{n}}_{i,0}$ represents the position of the same joint but before training. This regularization ensures that the motion graph, which is initialized close to the object geometry prior training, does not drift away from the object during optimization.

\noindent\textbf{2D Keypoints Regularization.} We project the 3D joints of the human kinematic tree onto the image plane and minimize the L1 distance between the projected joint positions and the 2D human keypoints predicted by SAPIENS. For our robotic experiments with kinematic trees on non-humanoid objects, see implementation details in Appendix~\ref{apn:robotics-details}.

\subsection{More Details of Real Robot Experiments}
\label{apn:robotics-details}


We evaluate our method with three real robot manipulation tasks, cloth folding, microwave door adjusting, and rope bending as shown in Fig.~\ref{fig:visual-plan}. We reconstruct the green cloth using a deformable graph, while applying kinematic trees for the microwave and rope. The canonical frame is set to the first frame for all tasks. 
We use SAM2 for object mask extraction, where we initialize the mask by clicking the object's central region in the first frame, then propagate it across the entire video. We capture the green cloth scene using three D415 RGB-D cameras, while the microwave and rope scenes are each recorded with a single D415 RGB-D camera. All the three dynamic objects are reconstructed from human demonstration videos.

For the green cloth, we initialize the motion graph at \( t = 0 \) using a rectangular mesh generated via NanoMesh~\cite{nanomesh} (Fig.~\ref{fig:visual-plan}, first column). For the microwave and rope---which lack standardized keypoint structures---we manually define their kinematic trees. The microwave is represented as a 2-link graph, with 3D keypoints annotated in the first (\( t = 0 \)) and final (\( t = T-1 \)) frames. 
For the rope, we annotate ten 2D keypoints at the first frame, and track them across frames using CoTracker~\cite{karaev2024cotracker3}.
The rope's kinematic tree is initialized by projecting the ten manually annotated 2D keypoints at the first frame into 3D space using depth information, with the central keypoint serving as the root. During optimization, we apply 2D keypoint regularization to the rope but omit it for the cloth and microwave.  

Given a test scene, we reconstruct its motion graph by optimizing the motion graph parameters of our learned dynamic Gaussian model. The optimization minimizes the L1 distance between the rendered image and the test scene image, while keeping the Gaussian parameters fixed. This process typically converges within a minute, and is akin to pose estimation for rigid objects but uses motion graphs which generalized to more flexible structures.


Given the reconstructed motion graph of a test scene, we simulate object and the corresponding end-effector movements using object-specific geometric heuristics.  For cloth, we simulate diagonal folding by sampling random bending axes and angles; for the microwave, we adjust the relative rotation angle between its two rigid links; and for the rope, we bend its single-chain structure at a randomly selected joint with a sampled angle.
Despite these simple heuristics, it is possible to integrate the motion graph simulation with a more powerful physics engine, as a future direction that we discussed in Sec.~\ref{sec:limitation}. At each timestep, we evaluate candidate motions by comparing their rendered images against the goal image, then select the motion that maximizes the PSNR. For each task, we evaluate performance across ten different goal images, each selected to be achievable through folding or door-adjustment actions. Our method reliably solve the cloth folding and microwave door manipulation tasks (10/10), while rope bending attains a 70\% success rate (7/10). We illustrate and discuss a typical failure case in Fig.~\ref{fig:limitation-robot}. Fig.~\ref{fig:kuka-robot} demonstrates our Kuka robot performing all three tasks: cloth folding, microwave door adjustment, and rope bending.

\subsection{Ablation Studies}
\label{sec:e:ablation}
\begin{figure}[t]
    \centering
    \includegraphics[width=0.7\linewidth]{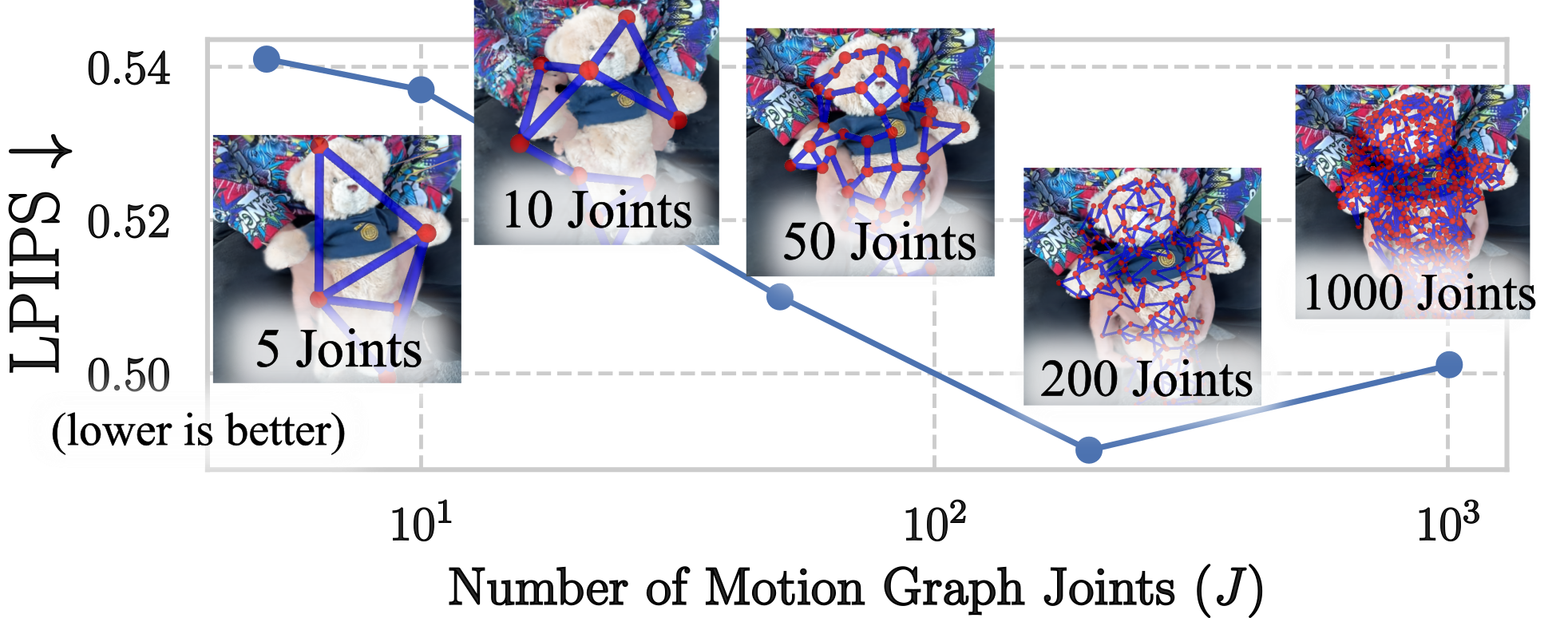}
    \caption{Relationship between novel view rendering quality and motion graph size.}
    \label{fig:graph_size}
\end{figure}

\begin{figure}[t]
    \centering
    \includegraphics[width=0.9\linewidth]{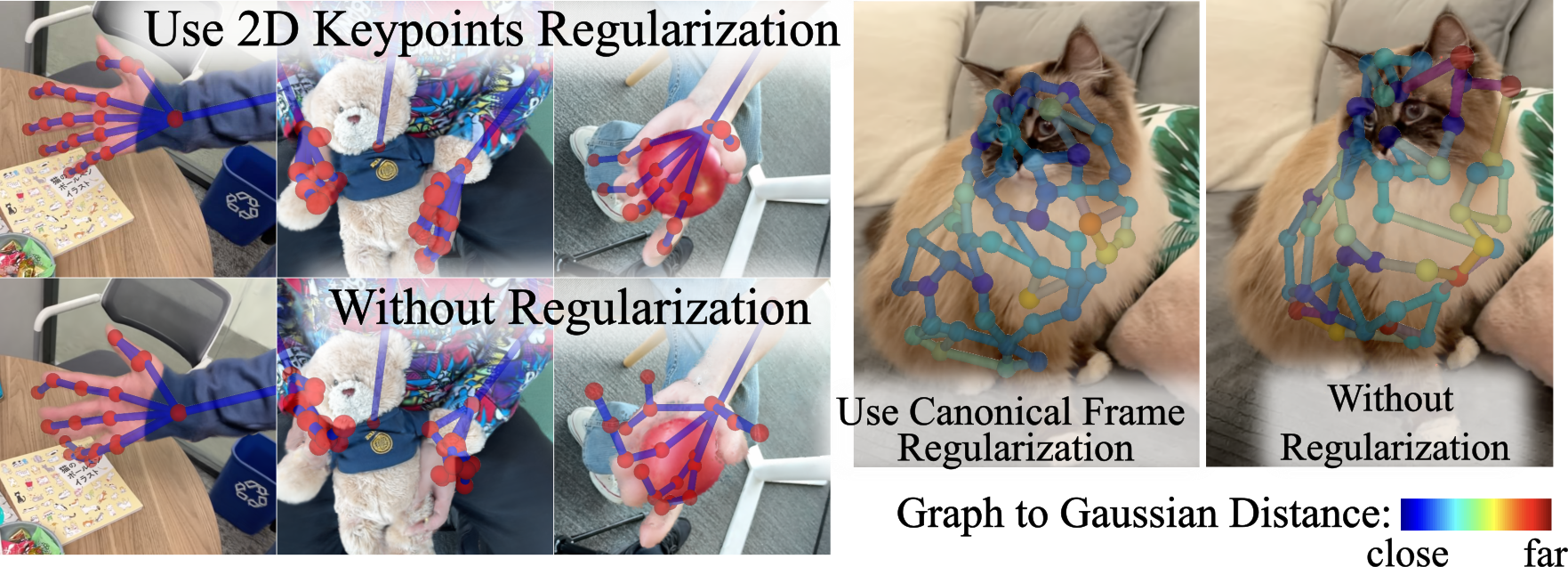}
    \caption{Visualization of rendered images and motion graphs with and without 2D human keypoint regularization (left) and canonical frame regularization (right).}
    \label{fig:regularization}
\end{figure}

\noindent\textbf{Motion Graph Size.} Fig.~\ref{fig:graph_size} illustrates the relationship between novel view rendering quality and the number of joints in a motion graph. We observe that the best rendering quality for the teddy scene is achieved with a motion graph of 200 joints. However, increasing the number of joints to 1000 leads to a noticeable drop in quality, suggesting the existence of an optimal graph size. This finding highlights the potential of learning adaptive graph structures that dynamically adjust to different objects.

\vspace{0.5em}\noindent\textbf{2D Keypoints Regularization.} Fig.~\ref{fig:regularization} compares rendered images and motion graphs with and without 2D human keypoint regularization. We observe that enforcing consistency between the projected 3D kinematic tree and 2D human keypoints (detected by SAPIENS~\cite{khirodkar2024sapiens}) improves performance, reducing LPIPS by 0.01 to 0.02. More importantly, this regularization produces cleaner and more meaningful motion graphs that align well with the human 3D structure. Without 2D keypoint regularization, the motion graph tends to drift or produce suboptimal structures. For example, if two fingers move together, the kinematic tree may incorrectly control both with a single finger link, leaving other finger links static. This highlights the importance of 2D keypoint regularization in learning accurate and interpretable 3D motion graphs.

\vspace{0.5em}\noindent\textbf{Canonical Frame Regularization.} The right side of Fig.~\ref{fig:regularization} compares rendered images and motion graphs with and without canonical frame regularization, which minimizes the distance between the learned motion graph's joint positions and their initial positions at the canonical frame. The joints and links of the motion graph are color-coded based on their distance to the Gaussians on the object surface. With canonical frame regularization, the motion graph remains closely aligned with the object surface. Without it, however, the graph joints become spiky (indicated by red-colored joints) and drift outward, detaching from the object.


\begin{figure}[t]
    \centering
    \includegraphics[width=\linewidth]{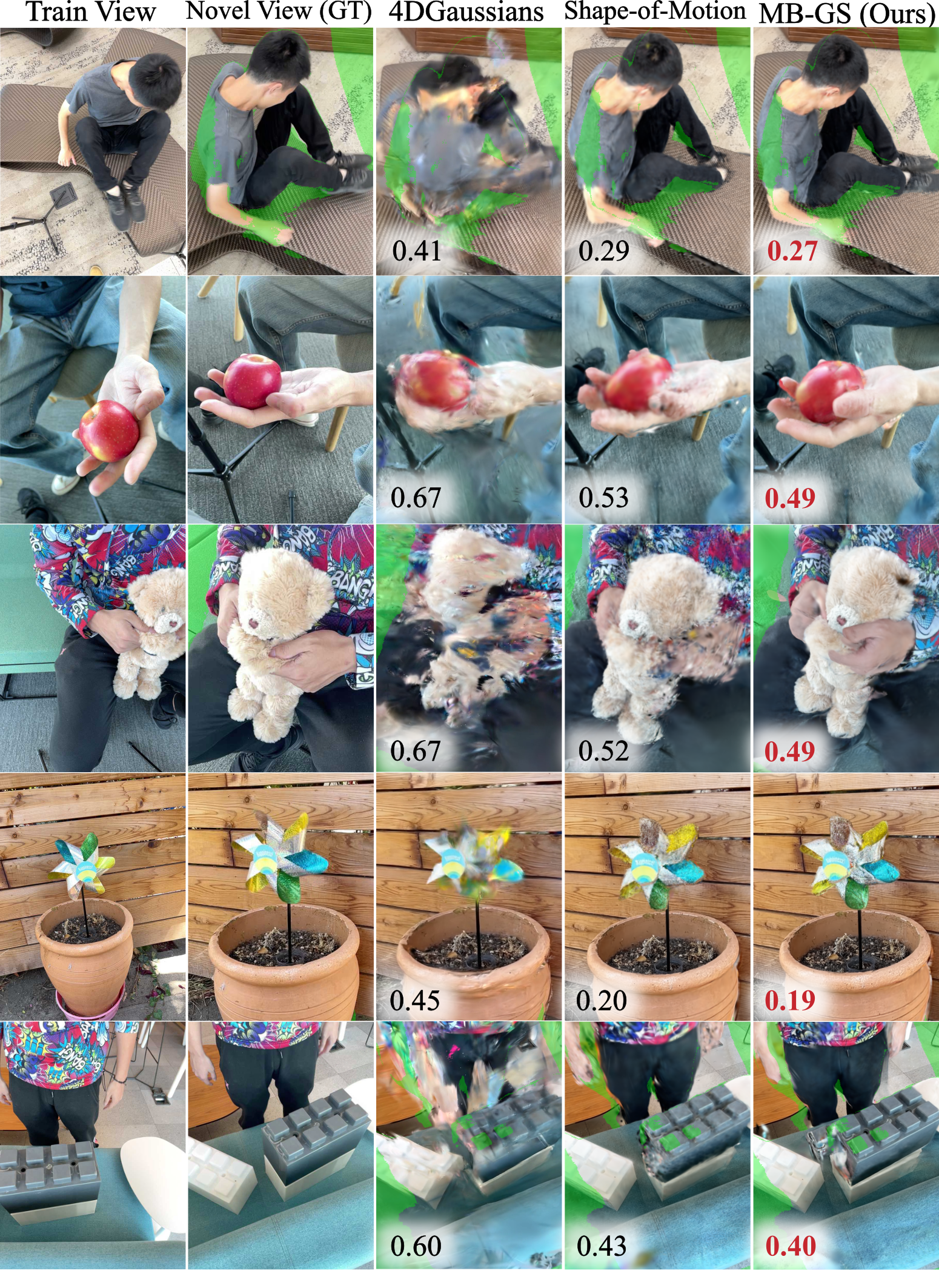}
    \caption{Visualization of novel view rendering on iPhone dataset compared with other methods. Regions in green are excluded from evaluation, as these pixels are never seen in training videos. LPIPS on bottom left with best colored \textcolor{darkred}{red}. Our method renders sharper, more complete, and perceptually higher-quality novel views.}
    \label{fig:quality-compare}
\end{figure}

\begin{figure}[t]
    \centering
    \includegraphics[width=0.9\linewidth]{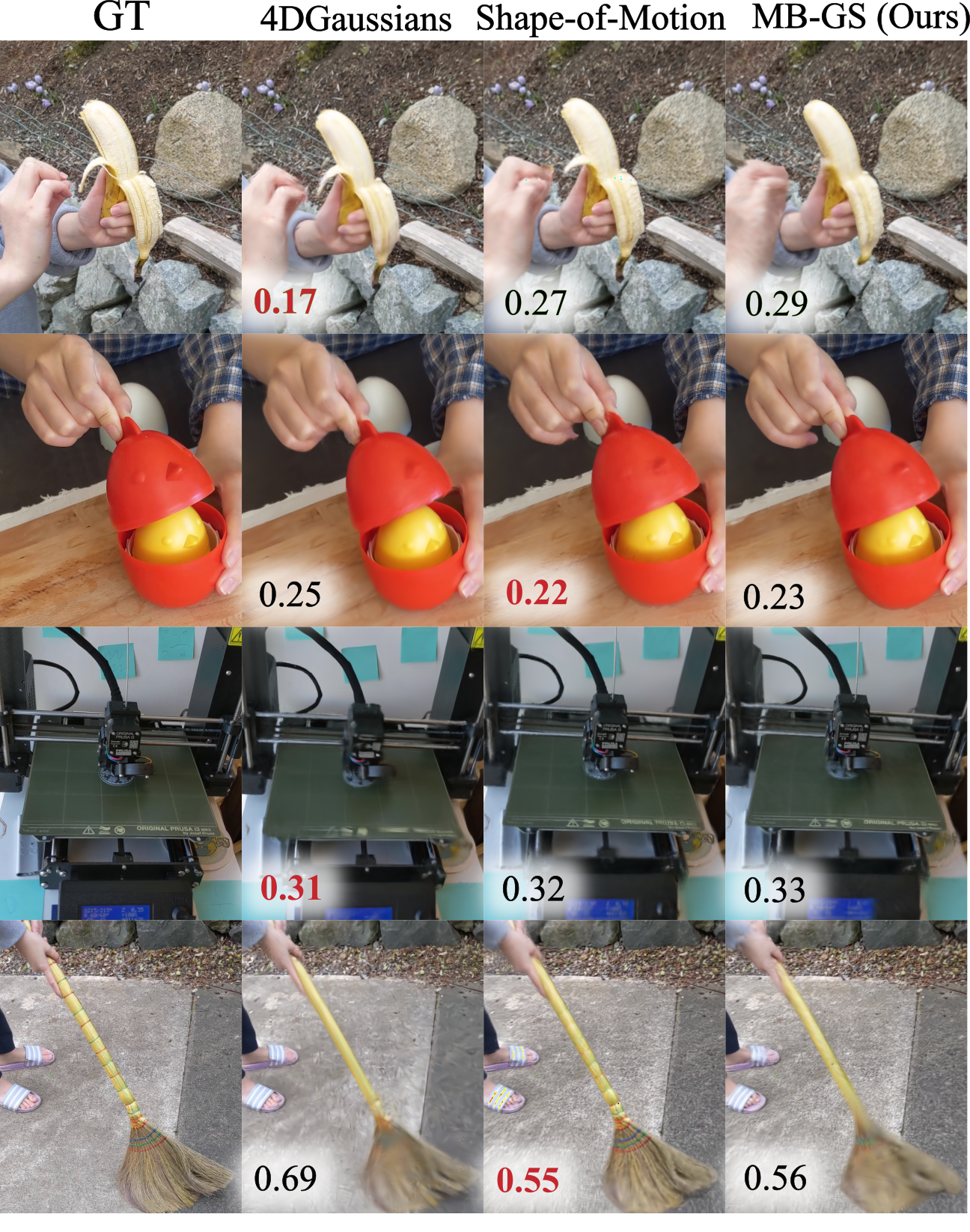}
    \caption{Visualization of results in HyperNerf dataset compared with other methods.  LPIPS on bottom left with best colored \textcolor{darkred}{red}. Our method achieves comparable rendering quality and LPIPS with state-of-the-art on  the chicken, 3D printer, and broom scenes (rows 2 to 4). On the peel-banana scene, our method produces lower-quality results.}
    \label{fig:hypernerf}
\end{figure}

\end{document}